\documentclass[12pt]{article}
\usepackage[left = 1 in, right = 1 in, top = 1 in, bottom = 1 in]{geometry}              
\usepackage{graphicx,wrapfig,lipsum,comment}
\usepackage{amssymb,amsmath,amsthm,amsfonts,bbm,bm,mathrsfs,xcolor}
\usepackage{mhequ}
\usepackage{mathabx}
\usepackage[numbers]{natbib}
\usepackage[colorlinks,citecolor=blue,urlcolor=blue,filecolor=blue,backref=page]{hyperref}
\usepackage{accents,url}
\usepackage[colorinlistoftodos,prependcaption,textsize=tiny]{todonotes}

\usepackage{subfig}

\usepackage{algorithm2e}

\def\m{\mathcal}
\def\mb{\mathbb}

\def\T{{\mathrm{\scriptscriptstyle T} }}

\newcommand{\bld}[1]{\mbox{\boldmath $#1$}}

\newcommand{\blds}[1]{\mbox{\scriptsize \boldmath $#1$}}

\newcommand{\be}{\begin{equs}}
\newcommand{\ee}{\end{equs}}

\numberwithin{equation}{section}
\theoremstyle{plain}
\newtheorem{theorem}{Theorem}

\newtheorem{definition}{Definition}

\newcommand{\beginsupplement}{%
        \setcounter{lemma}{0}
        \renewcommand{\thelemma}{S\arabic{lemma}}%
        \setcounter{table}{0}
        \renewcommand{\thetable}{S\arabic{table}}%
        \setcounter{figure}{0}
        \renewcommand{\thefigure}{S\arabic{figure}}%
        \setcounter{section}{0}
        \renewcommand{\thesection}{S\arabic{section}}%
        \setcounter{equation}{0}
        \renewcommand{\theequation}{S.\arabic{section}.\arabic{equation}}%
        \setcounter{page}{1}
}

\title{Fair Clustering via Hierarchical Fair-Dirichlet Process}

\author{Abhisek Chakraborty, Anirban Bhattacharya, Debdeep Pati
}

\date{Department of Statistics, Texas A\&M University,
College Station, TX, USA\\
May 24, 2023}

\begin{document}
\maketitle

\begin{abstract}
    The advent of ML-driven decision-making and policy formation has led to an increasing focus on algorithmic fairness. As clustering is one of the most commonly used unsupervised machine learning approaches, there has naturally been a proliferation of literature on {\em fair clustering}. A popular notion of fairness in clustering mandates the
clusters to be {\em balanced}, i.e., each level of a protected attribute must be approximately equally represented in each cluster. Building upon the original framework developed in  \citet{NIPS2017_978fce5b}, this literature has rapidly expanded in various aspects. In this article, we offer a novel model-based formulation of fair clustering, complementing the existing literature
which is almost exclusively based on optimizing appropriate objective functions. We first rigorously define a notion of fair clustering in the population level under a model mis-specified framework, with minimal assumptions on the data-generating mechanism. We then specify a
Bayesian model equipped with a novel hierarchical prior specification to encode the notion of balance in resulting clusters, and whose posterior targets this population-level object. A carefully developed collapsed Gibbs sampler ensures efficient computation, with a key
ingredient being a novel scheme for non-uniform sampling from the space of binary matrices with fixed margins, utilizing techniques from optimal transport towards constructing proposals.
Impressive empirical success of the proposed methodology is demonstrated across varied numerical experiments, and benchmark data sets. Importantly, the benefits of our approach are not merely limited to the specific model we propose -- thinking from a generative modeling
perspective allows us to provide concrete guidelines for prior calibration that ensures desired distribution of balance {\em a-priori}, develop a concrete notion of optimal recovery in the fair
clustering problem, and device schemes for principled performance evaluations of algorithms.
\end{abstract}


\section{Introduction}
Algorithmic decision-making is increasingly  employed in critical aspects affecting human lives, e.g credit, employment, education, criminal justice;  and hence fairness in machine learning \citep{https://doi.org/10.48550/arxiv.1104.3913, 10.1145/3494672, doi:10.1146/annurev-statistics-042720-125902} has emerged as a primary pillar of recent research focus. Discrimination refers to unfavourable treatment of entities due to the membership to certain demographic groups that are determined by factors referred to as \emph{protected attributes}. Hence, the goal of group fairness is to design algorithms that make fair decisions devoid of discrimination due to membership to label of a protected attribute. Noticeably, in the nascent stages of the quest for fairness in machine learning algorithms,  much of the literature were targeted towards supervised learning techniques and an  obvious need was felt to consolidate the notions of fairness in the context of unsupervised learning problems, such as clustering. \citet{NIPS2017_978fce5b} introduced the concept of \emph{balance} as a criterion for fair clustering; defining a clustering mechanism to be fair if the resulting clusters share a common ratio of data points representing individuals belonging to the different groups of the protected attribute; and  discussed both the $k$-center and the $k$-median problems with two labels of the protected attribute, i.e the \emph{two-color} case. Subsequently, a number of follow up works have attempted to extend this to a \emph{multi-color} case \citep{https://doi.org/10.48550/arxiv.2002.07892}, where the protected attribute has more than two labels.  \citet{NEURIPS2020_95f2b84d} further generalized these by assuming imperfect knowledge of group membership through probabilistic assignments.  \citet{NEURIPS2019_fc192b0c} extended the scope by allowing the user to specify parameters that controls the extent of fair representation, general $\mb{L}_p$ norm of the clustering objective, and by considering the case where individuals can lie in multiple protected groups.  At the same time, the mechanisms for ensuring fairness in various other avatars of clustering, e.g spectral clustering  \citep{https://doi.org/10.48550/arxiv.1901.08668}, correlation clustering \citep{DBLP:journals/corr/abs-2002-02274}, hierarchical clustering \citep{DBLP:journals/corr/abs-2006-10221}, have emerged. Other notions of fairness in the context of clustering have been looked at too, e.g individual fairness \citep{https://doi.org/10.48550/arxiv.2006.04960, https://doi.org/10.48550/arxiv.2002.06742, DBLP:journals/corr/abs-2106-12150}, proportional fairness \citep{DBLP:journals/corr/abs-1905-03674}.  Perhaps unsurprisingly, fairness in clustering has been studied in conjunction with other pressing aspects of modern machine learning as well, e.g privacy \citep{DBLP:journals/corr/abs-1802-02497}, data summarization \citep{https://doi.org/10.48550/arxiv.1901.08628}, robustness \citep{DBLP:journals/corr/abs-1907-08906}, to name a few. For a comprehensive review of the literature, interested readers are encouraged to browse this website on \href{https://www.fairclustering.com/}{\textcolor{purple}{fair-clustering}}, and refer to the recent comprehensive review article \citet{https://doi.org/10.48550/arxiv.1104.3913}.

\subsection*{Our Contributions}
\begin{itemize}
    \item \textbf{(1)} We take a novel model-based approach to tackle the problem of clustering under balance constraints. Thinking from a generative modeling perspective allows us to develop a concrete notion of \emph{optimal recovery} in this problem, and subsequently device a scheme for principled \emph{performance evaluation} of algorithms. 
    
    \item \textbf{(2)} We specify a hierarchical model equipped with a novel prior specification, termed as hierarchical fair Dirichlet process, to  encode the notion of balance in resulting clusters;  provide concrete guidelines for the prior calibration that ensures a desired distribution of \emph{balance} a-priori; and develop a collapsed Gibbs sampler that exploits a carefully designed gamma prior specification on the shared concentration parameter ensuring rapid mixing and scalability.
    
    \item  \textbf{(3)} The most prominent computational bottleneck in our approach involving sampling cluster membership indices subject to balance constraints, which in turn poses a difficult non-uniform sampling problem in the space of binary matrices with fixed margins. This is carefully navigated via proposing a novel weighted rectangular loop scheme equipped with an integer-valued optimal-transport based proposal mechanism. This scheme may be of independent importance in many applications in  neurophysiology, sociology, psychometrics and ecology \citep{BRUALDI198033, curveball, recloop}.
    
    \item \textbf{(4)} Our approach enjoys seamless adaption  to include key extensions present in  the fair clustering literature \citep{NIPS2017_978fce5b, https://doi.org/10.48550/arxiv.2002.07892, NEURIPS2020_95f2b84d}; and inherits the usual benefits of model based clustering, e.g,  simultaneous learning of the number of clusters and the cluster memberships \citep{doi:10.1198/016214502760047131, 10.1093/biomet/asac051, doi:10.1080/01621459.2016.1255636};
providing a framework to potentially handle an enormous range of pressing practical  complexities --  non-isotropic covariances within clusters, mixed data type, censoring, missing values, variable selection \citep{complicated}, covariate adjustment, etc.
    
\end{itemize}

\section{Bayesian Clustering with Fairness Constraints}
\subsection{Probabilistic Framework}\label{ssec:principles}
Let $\mb Z_{\ge 0}$ (resp. $\mb R_{\ge 0}$) denote the set of non-negative integers (reals). For a positive integer $t$, denote $[t] :\,= \{1, \ldots, t\}$, and let $\Delta^{t-1} \subset \mb R^t$ denote the $(t-1)$-dimensional probability simplex, i.e., $\Delta^{t-1} = \{x \in \mb R_{\ge 0}^t \,:\, \langle 1_t, x \rangle = 1\}$. For two probability distributions $p_1, p_2$ with same support, the Kullback--Leibler divergence from $p_1$ to $p_2$ is $\mbox{KL}(p_1\,||\,p_2) = E_{p_1} \log(p_1/p_2)$; and the Symmetrised Kullback-Liebler divergence between $p_1$ and $p_2$ is $\mbox{KL}_{\rm sym}(p_1, p_2) = \mbox{KL}(p_1\,||\,p_2) + \mbox{KL}(p_2\,||\,p_1)$. A $t$-dimensional random vector $u$ follows a Dirichlet distribution with concentration parameter $\bld \alpha \in \mb R_+^t$, denoted 
$\bld u\sim\mbox{Dir}(\bld \alpha)$, if $u_i = v_i/\sum_{j=1}^t v_j$ with $v_i\overset{ind.}\sim\mbox{Gamma}(\alpha_i, 1)$ for $i \in [t]$.

Suppose we observe data $\{(\bld x_i, a_i)\}_{i=1}^N$, where $\bld x_i$ denotes the $d$-variate observation for the $i$-th data unit, and $a_i$ the label of the protected attribute. For each $a$, let $\{\bld x_i^{(a)}\}_{i=1}^{N_a}$ denote the observations corresponding to the $a$-th level of the protected attribute, where $N_a = \sum_{i=1}^N \mathbf{1}(a_i = a)$ and  $\sum_{a=1}^r N_a = N$. The goal of fair clustering is to assign the data points $\{(\bld x_i, a_i)\}_{i=1}^N$ into clusters $\bld C = (C_1,\ldots,C_K)$, $\dot{\bigcup} C_k = [N]$, respecting the notion of balance  \citep{NIPS2017_978fce5b}, presented next.

\begin{definition} [Balance, \citep{NIPS2017_978fce5b}]
Given  $\{(\bld x_i, a_i) \in \m X \times \m A, \ i =1\ldots, N\}$ such that $a_i = a$ for $i= \sum_{j=1}^{a-1} N_j + 1,\ldots, \sum_{j=1}^{a} N_j$ where $a\in [r]$ and $N_0$ = 0, the balance in $C_k$ is defined as $\mbox{Balance}(C_k) = \min_{1\leq j_1 < j_2 \leq r} \big\{|C_{kj_1}|/|C_{kj_2}|, |C_{kj_2}|/|C_{kj_1}|\big\}$
where $|C_{kj}|$ denote the number of observations in $C_k$ with $a = j$. The overall balance of the clustering is  $\mbox{Balance}(\bld C) = \min_{k=1,\ldots,K} \mbox{Balance}(C_k)$.
The higher this measure is for a clustering configuration, the fairer is the clustering.     
\end{definition}

With the above definition of balance, \citet{NIPS2017_978fce5b} introduced the concept of \emph{fairlets} that are minimal fair sets which approximately preserve the clustering objective of choice; and demonstrated  that any fair clustering problem can be reduced to first finding a fairlet decomposition of the data via solving a \emph{minimum cost flow} problem, and then resorting to classical  clustering algorithms, e.g k-means, k-center. While the follow-up works \citep{ https://doi.org/10.48550/arxiv.2002.07892, NEURIPS2020_95f2b84d, DBLP:journals/corr/abs-1802-02497, https://doi.org/10.48550/arxiv.1901.08628} greatly increased the scope of fair clustering from various aspects, a precise statistical framework is largely missing  in the literature, which  in turn makes it difficult to define the notion of the \emph{true fair clustering configuration} at the \emph{population level}, theoretically study optimal discovery/clustering consistency under fairness constraints, and device principled performance measures to compare  algorithms. This genuinely ordains a concrete probabilistic treatment to study clustering with balance constraints, and the following discussion precisely targets to achieve this. 

We assume that $\{(x_i, a_i)\}_{i=1}^N$ are independent copies of the random vector $(X, A)\in\m X\times A\subset \mb{R}^d\times \m A$;  $\m A = [r]$ without loss of generality. A weight vector $\bld\xi\in\Delta^{r-1}$ records the population proportions of  the different labels of the protected attribute, which is assumed to be completely known. The generative mechanism for $(X, A)$ is hypothesised to be governed by $ A\sim \mathcal{P}^{\star}_{A}\equiv \mbox{Multinomial}(1, \bld\xi),\ (X, Z)\mid A  \sim \mathcal{P}^{\star}_{X, Z\mid A}\equiv \mathcal{P}^{\star}_{X\mid Z, A}\times \mathcal{P}^{\star}_{Z\mid A}$, where the probability measure $\mathcal{P}^{\star}_{X, A, Z}$ is unknown and  the random variable $Z$  is the latent/unobserved clustering index.  In practice, we shall only observe independent copies of  $(X, A)$; and the goal is to learn the marginal generative mechanism $\mathcal{P}^{\star}_{Z}$ of the clustering index $Z$, such that the clustering  is \emph{balanced}.  To consolidate the notion of \emph{balance} at population level, we first introduce the space of joint probability measures of $(X, Z, A)$ denoted by $\mathbb{P}$; and then define a subset $\mathbb{P}^{(R)} = \big\{\mathcal{P}_{X, Z, A}\in\mathbb{P}\ :\ \mathcal{P}_{A\mid  Z} = \mathcal{P}_{A} \big\}\subset \mathbb{P}$ that collects all possible generative models under which every label of the protected attribute are  equally likely to appear within each cluster. However,  aiming for exact balance in the clusters may not be desired in practice, and in order to provide additional flexibility in specifying the notion of balance, we introduce 
$\mathbb{P}^{(R)}_{\varepsilon} = \big\{\mathcal{P}_{X, Z, A}\in\mathbb{P}\ :\ \mbox{KL}(\mathcal{P}_{A}\times  \mathcal{P}_{Z}\  ||\ \mathcal{P}_{A, Z})\leq \varepsilon \big\}$.
The user-defined quantity  $\varepsilon\geq 0$  controls the extent of departure from balance, and  we say that  $(A, Z)$ satisfies $\varepsilon$-\emph{balance} under $\mathcal{P}_{X, Z, A}$  iff
$\mbox{KL}(\mathcal{P}_{A}\times  \mathcal{P}_{Z}\  ||\ \mathcal{P}_{A, Z})\leq \varepsilon$.
By definition, we have $\mathbb{P}^{(R)}_{0} = \mathbb{P}^{(R)}$, and it refers to the class of generative mechanisms where strict balance is maintained. The quantity at the heart of our definition of balance, $\mbox{KL}(\mathcal{P}_{A}\times  \mathcal{P}_{Z}\  ||\ \mathcal{P}_{A, Z})$, is the \emph{mutual information} \citep{10.5555/1146355} between $(A, Z)$, and is indeed a key pivot to test for departure of the joint distribution of $(A, Z)$ from the independence table \citep{10.1093/biomet/asz024, AI2022}. Noteworthy, for a fixed $\varepsilon$, the true generative model $\mathcal{P}^{\star}$ may not belong to $\mathbb{P}^{(R)}_{\varepsilon}$. However, since we wish to ensure that our clustering procedure is $\varepsilon$-\emph{balanced}, the inferential goal constitutes finding the ``best" approximation of the true generative model $\mathcal{P}^{\star}$ within the restricted class $\mathbb{P}^{(R)}_{\varepsilon}$. Readers may associate this framework to that of  maximum likelihood estimation under model misspecification \citep{10.2307/1912526}; or  variational inference \citep{doi:10.1080/01621459.2017.1285773}, where we intend to find the ``best" approximation of an unknown probability distribution  within a well-behaved variational family. We appeal to the extensive literature on  these topics, in order to ensure principled treatment of our inferential goal. To that end, two definitions are in order. 

\begin{definition}[Pseudo True Fair Clustering Distribution, PT-FCD]
 For fixed $\varepsilon\geq 0$, the Pseudo-true clustering distribution restricted to $\mathbb{P}^{(R)}_{\varepsilon}$ is defined as, 
\begin{equation}\label{def:map_fcd}
     \mathcal{P}^{\varepsilon}_{Z_{\rm PT-FCD}}= \mbox{\rm argmin}_{\mathcal{P}_{A, Z, X}\in \mathbb{P}^{(R)}_{\varepsilon}} \mbox{\rm KL}(\mathcal{P}^{\star}_{Z\mid X, A}\ ||\ \mathcal{P}_{Z\mid X, A}),
\end{equation}
where $\mathcal{P}_{Z\mid X, A}$ denotes the conditional distribution of $Z\mid X, A$ obtained on marginalization from a $\mathcal{P}_{A, Z, X}\in \mathbb{P}^{(R)}_{\varepsilon}$. Further, the probability distribution of the number of unique elements induced from $\mathcal{P}^{\varepsilon}_{Z_{\rm PT-FCD}}$ in Definition \ref{def:map_fcd} is referred to as the \emph{pseudo-true distribution of number of clusters}.   
\end{definition}

Note $\mathcal{P}^{\varepsilon}_{Z_{\rm PT-FCD}}$ represents the \emph{KL projection} of the true $\mathcal{P}^{\star}_{Z\mid X, A}$ within the $\varepsilon$-\emph{balanced} class of generative models, and the minimization problem in \eqref{def:map_fcd} can be recasted as 
\begin{align*}
 \mbox{\rm max}_{\mathcal{P}_{A, Z, X}\in \mathbb{P}} \big[\mathbb{E}_{\mathcal{P}^{\star}_{Z\mid X, A}}( \log\mathcal{P}_{Z\mid X, A}) -\lambda_{\varepsilon} \mbox{KL}(\mathcal{P}_{A}\times  \mathcal{P}_{Z}\  ||\ \mathcal{P}_{A, Z})\big], \end{align*}
where $\lambda_{\varepsilon}\geq 0$ is a Lagrange's multiplier that depends on $\varepsilon$. This dual formulation would  be crucial in the proof of Theorem \ref{th1} below.

\subsection{Restricted Posterior Maximization}\label{subsec:res_post_max}
In this section, we develop the general template of a data-driven procedure that targets to optimally recover the population level quantity $\mathcal{P}^{\varepsilon}_{Z_{\rm PT-FCD}}$. To keep parity with sub-section \ref{ssec:principles}, suppose  $\mathbb{Z}$ is the collection of all possible clustering configurations, and $\mathbb{Z}^{(fair)}$ is the subset of $\mathbb{Z}$ where \emph{balance} is maintained. That is, $\mathbb{Z}^{(fair)} = \big\{\bld z\in \mathbb{Z}:  \mathcal{C}_{\blds z} \text{ is balanced}\big\}$,
where $\mathcal{C}_{\blds z}$ is the clustering configuration induced by ${\bld Z = \bld z}$. A related relaxed notion of $\mathbb{Z}^{(fair)}$ constitutes defining $\mathbb{Z}^{(fair)}(\varepsilon) = \big\{\bld z\in \mathbb{Z}:\mbox{KL}(\mathcal{P}_{\blds A}\times\mathcal{P}_{\blds Z}\ ||\ \mathcal{P}_{\blds A, \blds Z})\leq \varepsilon\big\}$,
where  $\varepsilon\geq0$ is a  hyper-parameter, and we  have $\mathbb{Z}^{(fair)}(0)= \mathbb{Z}^{(fair)}$. 

\begin{definition}[Maximum-a-Posteriori fair cluster, MAP-FC]
The modal clustering configuration obtained from the conditional distribution of $\bld Z\mid \bld X, \bld A$ restricted to $\mathbb{Z}^{(fair)}(\varepsilon)$, denoted by $\mathcal{P}^{(fair)}_{\blds Z\mid\blds X, \blds A}(\bld z\mid\bld x, \bld a)\equiv\mathcal{P}_{\blds Z\mid\blds X,\blds A}(\bld z\mid\bld x, \bld a)\times \mathbf{1}_{\mathbb{Z}^{(fair)}(\varepsilon)}(\bld z)$, is termed as the Maximum-a-Posteriori fair clustering configuration, i.e, 
\begin{equation}\label{def:map_fc} 
     {\bld z}^{\varepsilon}_{\rm MAP-FC}= \mbox{\rm argmax}_{\blds z\in  \mathbb{Z}^{(fair)}(\varepsilon)} \mathcal{P}_{\blds Z\mid\blds X,\blds A}(\bld z\mid\bld x, \bld a).
\end{equation}    
\end{definition}
The maximization problem in \eqref{def:map_fc} can be expressed equivalently as $\mbox{\rm argmax}_{\blds z\in  \mathbb{Z}(\varepsilon)} \mathcal{P}_{\blds Z\mid\blds X,\blds A}(\bld z\mid\bld x, \bld a) - \lambda_{\varepsilon}\mbox{KL}(\mathcal{P}_{\blds A}\times\mathcal{P}_{\blds Z}\ ||\ \mathcal{P}_{\blds A, \blds Z})$, where $\lambda_{\varepsilon}$ is the Lagrange's multiplier.
Further, the number of unique elements in $\bld z^{\varepsilon}_{\rm MAP-FC}$ in Definition \ref{def:map_fc} is referred to as the \emph{Maximum-a-Posteriori number of clusters}. In principle, we can potentially target  $z^{\varepsilon}_{\rm MAP-FC}$ via a two-staged  post-processing scheme, where we first fit a flexible Bayesian model and obtain samples from the unconstrained posterior of $\bld Z\mid\bld X$. Then, as a post-processing step, we sort the posterior samples according to their posterior probability and report the clustering configuration with the highest posterior probability while satisfying the balance constraint. However, such an approach is  clearly computationally prohibitive, and we provide a novel get-around via proposing a  HFDP prior in Section \ref{sec:model_prior} that encodes the notion of approximate balance by shrinking towards $\mathbb{Z}^{(fair)}(\varepsilon)$, with the degree of shrinkage controlled by hyperparameters which can be appropriately calibrated. This entirely avoids the need for the aforesaid post-processing, as the posterior samples automatically encode fairness with high probability. 

Note that,  $\bld X\mid  \bld A$ assumes a mixture distribution of the form $\bld X\mid  \bld A \sim  \sum_{\blds z} \zeta_{\blds a,\blds z} \mathcal{P}_{\blds X\mid\blds Z,\blds A}$, where $\mathcal{P}_{\blds Z\mid \blds A}(\bld Z=\bld z\mid \bld A =\bld a) = \zeta_{\blds a,\blds z}$ and $\sum_{\blds z}\zeta_{\blds a,\blds z} = 1\ \forall\ \bld a\in[r]^N$. Then, with a slight abuse of notation, we have
$    \mathcal{P}^{(fair)}_{\blds Z\mid\blds  X, \blds A}(\bld z\mid\bld x, \bld a)=  [  \zeta_{\blds a,\blds z} \mathcal{P}_{\blds X\mid \blds Z, \blds A}(\bld x\mid\bld z,\bld a) \mathcal{P}_{\blds A}(\bld a)\times \mathbf{1}_{\mathbb{Z}^{(fair)}(\varepsilon)}(\bld z)]/\mathcal{G}_N$,
where $\mathcal{G}_N = \sum_{\blds z} [\zeta_{\blds a,\blds z} \mathcal{P}_{\blds X\mid \blds Z, \blds A}(\bld x\mid \bld z,\bld a) \mathcal{P}_{\blds A}(\bld a)]\times \mathbf{1}_{\mathbb{Z}^{(fair)}(\varepsilon)}(\bld z)$. 
Given a cluster configuration $\hat{\bld z}$, define
\begin{equation}\label{criterion}
\mbox{Fair-Score}(\hat{\bld z}) :\, =
\log\big[  \zeta_{\bld a,\hat{\bld z}} \mathcal{P}_{\blds X\mid \blds Z, \blds A}(\bld x\mid \hat{\bld z},\bld a) \mathcal{P}_{\blds A}(\bld a)]\times \mathbf{1}_{\mathbb{Z}^{(fair)}(\varepsilon)}(\hat{\bld z})\big]  
\end{equation}
Clearly, Fair-Score as defined above is maximized at ${\bld z}^{\varepsilon}_{\rm MAP-FC}$. An empirical version, replacing population quantities by data-driven estimates, can be used to compare different clustering methods. The Fair-Score thus serves as a criterion towards a principled assessment of the clustering performance as it strikes a balance between posterior maximization while maintaining fairness. For a flavor, refer to Figure \ref{fig:fair_score_validity} which illustrates that the modal clustering obtained from the HFDP procedure proposed in Section\ref{sec:model_prior} achieves higher fair-score compared to existing methods \citep{NIPS2017_978fce5b, https://doi.org/10.48550/arxiv.2002.07892, NEURIPS2020_95f2b84d}. 
 

In summary, Definition \ref{def:map_fcd} introduces a notion of $\varepsilon$-balanced clustering at the population level under a model mis-specified framework, with minimal assumptions on the data-generating mechanism. In parallel, having observed samples $\{(\bld x_i, a_i)\}_{i=1}^N$, Definition \ref{def:map_fc} discusses a restricted posterior maximization framework that targets the population-level object, of which the HFDP introduced in Section \ref{sec:model_prior} is a specific example.  To reconcile the notions of clustering with balance constraints at population level  \eqref{def:map_fcd} and its sample counterpart \eqref{def:map_fc}, an asymptotic equivalence result follows. To that, let us define a map $\mathcal{R}:[K]^N\to[0, 1]^K$ that takes a clustering configuration $\bld z\in[K]^N$ as input and outputs
the relative proportion of cluster indices $(1/N)[\sum_{i=1}^N \mathbf{1}(z_i = 1),\ldots, \sum_{i=1}^N \mathbf{1}(z_i = K)]^{\T}$. The marginal posterior of $(\bld Z \mid \bld X, \bld A)$ along with the reparameterization $\bld z\to \mathcal{R}(\bld z)$, yields the marginal posterior of $(\mathcal{R}(\bld Z)\mid \bld X, \bld A)$, denoted by $P_{\mathcal{R}(\blds Z) \mid \blds X, \blds A}$.
\begin{theorem}[An equivalence result]\label{th1}
In Definitions \ref{def:map_fcd} and \ref{def:map_fc}, for any fixed $\varepsilon\geq 0$, under clustering consistency conditions,
\begin{align*}
 \lim_{N\to\infty} \bigg[ \frac{P_{\mathcal{R}(\blds Z) \mid \blds X, \blds A}(\mathcal{R}(\bld z^{\varepsilon}_{\rm MAP-FC})\mid \bld x,\bld a)}{\sum_{\blds z} P_{\mathcal{R}(\blds Z) \mid \blds X, \blds A}(\mathcal{R}(\bld z)\mid \bld x,\bld a)\mathbf{1}_{\mathbb{Z}^{(fair)}(\varepsilon)}(\bld z)}\bigg]
= \mathcal{P}^{\varepsilon}_{Z_{\rm PT-FCD}}.   
\end{align*}
\end{theorem}
We defer the proof to the \textcolor{purple}{supplementary material}, and discuss the key take-aways from the Theorem \ref{th1}. Firstly, the equivalence result demonstrates that, given data $\{(\bld x_i, a_i)\}_{i=1}^N$, we can carefully construct probability models such that the resulting modal $\varepsilon$-balanced clustering configuration accurately recovers the  Kullback--Liebler projection of the true  population clustering distribution within the $\varepsilon$-balanced class, as sample size diverges to $\infty$. Secondly, on the practical front, Theorem \ref{th1} justifies the utility of the quantity in \eqref{criterion} as a principled measure for clustering performance comparison. 

\section{Model and Prior Specification}\label{sec:model_prior}
 Let $\m Z_{N, K} = \{ \bld z = (z_1, \ldots, z_N) : z_i \in [K] \ \text{for all} \ i \in [N] \}$ denote the space of all clustering configurations of $N$ observations into $K$ clusters. Any $\bld m = (m_1, \ldots, m_K) \in \mb Z_{\ge 0}^K$ such that $\sum_{k=1}^K m_k = N$ will be called a a {\em cluster occupancy vector}. Given such $\bld m$, let 
$
\m Z_{N, K, \blds m} = \{z \in \m Z_{N, K} \,: \, \sum_{i=1}^N \mathbf{1}(z_i = k) = m_k, \ i \in [N]\}
$
denote all clustering configurations with cluster occupancy vector $\bld m$.  Note that, $\m Z_{N, K, \blds m}$ can be uniquely characterised by the space of $N\times K$ binary cluster membership matrices with fixed column-sum $\bld m$ and row-sum $\bld 1_{N}$. This is crucially exploited for posterior sampling; see around equation \eqref{eq:non_u}. Next, define a function $\mbox{rd}: \mb N \times \Delta^{t-1} \to \mb Z_{\ge 0}^t$, so that for a positive integer $n \in \mb N$ and a probability vector $\bld u\in\Delta^{t-1}$, $\bld v = \mbox{rd}(n, \bld u)$ is given by 
$v_i = \mbox{round}(n u_i)$ for $i\in[t-1]$, where ``$\mbox{round}$" denotes the rounding function to the nearest integer, and $v_t = n - \sum_{i=1}^{t-1} v_i \ge 0$. Clearly, $\langle 1_t, \mbox{rd}(n, \bld u) \rangle = n$ for any $\bld u$. We use this $\mbox{rd}(\cdot)$ function to create a novel prior on cluster occupancy vectors below. 

We now  detail pieces of a hierarchical model to carry out Bayesian clustering with fairness constraints. The lowest level hyperparameters of our model are $(K, g, b)$, where $K$ is an upper bound on the number of clusters, and $g, b$ are positive parameters. Given these hyperparameters, we sample a global weight vector $\bld \beta \in \Delta^{K-1}$ and a concentration parameter $\alpha_0 \in (0, \infty)$ as in \eqref{hdpmm_1} below. Next, given $\alpha_0$ and $\bld \beta$, we independently sample a weight vector $\bld w^{(a)}$ corresponding to each level of the attribute $a\in[r]$ from $\mbox{Dir}(\alpha_0 \, \bld \beta)$. 
\begin{figure}
\begin{center}
\includegraphics[width=14cm, height =2.5cm]{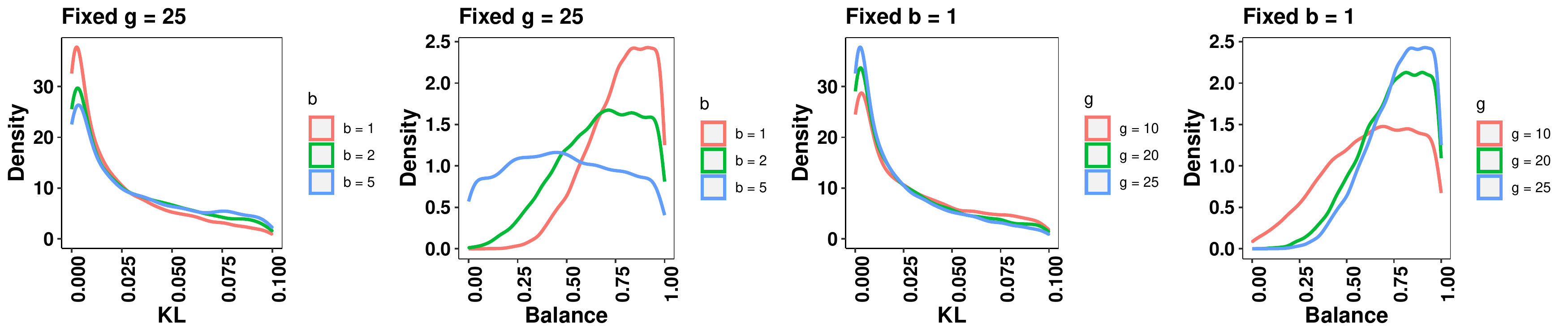} 
\caption{\emph{The first two plots, respectively, present the distribution of the $\mbox{KL}(w^{(1)}\ ||\ w^{(2)})$, and balance between $(\bld w^{(1)}, \bld w^{(2)})$, for varying $b$ with fixed $g$. The final two plots present the same quantities, now with varying $g$ with fixed $b$. In practice, we vary $(b,g)$ until we obtain the desired distribution of balance a-priori, and thus this provide a visual recipe for prior calibration. }}\label{fig:vary_b_g}
\end{center}
\end{figure}
\begin{align}\label{hdpmm_1}
\bld w^{(a)} \,|\, \alpha_0,  \bld \beta \stackrel{\textit{ind.}}{\sim}  \mbox{Dir}(\alpha_0 \, \bld \beta), \, a \in [r]; \
 \bld \beta \,|\, g \sim \mbox{Dir}(g/K,\ldots,g/K);\ \alpha_0 \,|\, g, b \ \sim\ \mbox{Gamma}(g,\ b). 
\end{align}
The concentration parameter $\alpha_0$ dictates how tightly the $\{\bld w^{(a)}\}_{a=1}^r$ concentrate around $\bld \beta$, and is critical in enabling a notion of balance in our model-prior specification. Figure \ref{fig:vary_b_g} provides a visual recipe for calibration of the prior on $\alpha_0$ with respect to the lowest level hyperparameters  $(g, b)$. In particular, given $(g, b)$ and $K = 2$, we obtain prior draws of $(\bld w^{(1)}, \bld w^{(2)})$ according to the equation block \eqref{hdpmm_1}, and present the distribution \emph{a-priori} of  $\mbox{KL}(w^{(1)}\ ||\ w^{(2)})$ and balance between $(\bld w^{(1)}, \bld w^{(2)})$. 

\begin{figure}[!h]
\begin{center}
\includegraphics[width=14.5cm, height =2.0cm]{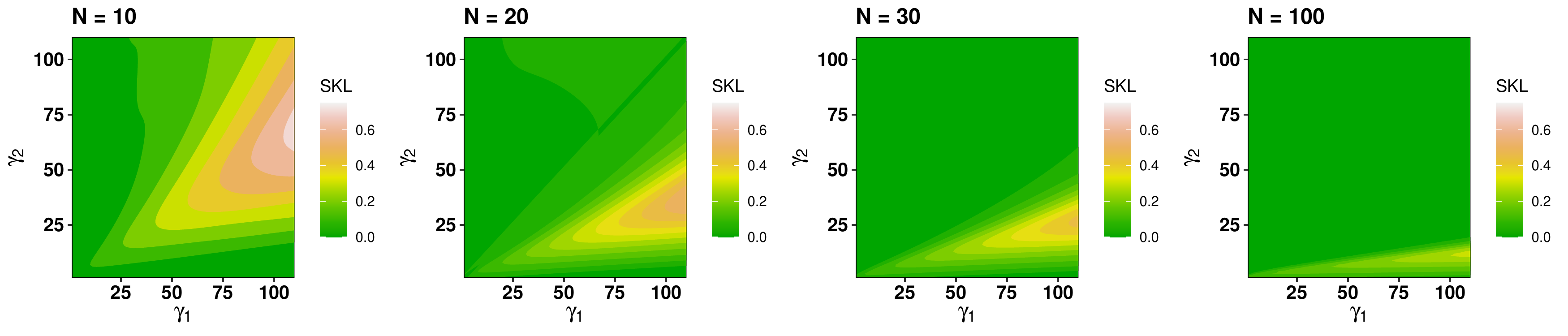} 
\centering{\caption{{\em Heatmap of the symmetrized KL distance between a lifted Beta distribution and a Beta-Binomial distribution against the Beta parameters $\gamma_1$ and $\gamma_2$. As $N$ increases, the symmetrized KL uniformly gets smaller, implying the distributions get closer over a wide range of parameter values.}}\label{plot:KL_Y1Y2}}
\end{center}
\end{figure}

Having drawn $\bld w^{(a)}$ for each label of the attribute, we draw the cluster configuration $\bld z^{(a)}$ for the $a$th sub-population in the following hierarchical manner to maintain strict control over the cluster sizes, 
\begin{align}\label{hdpmm_2}
\bld z^{(a)} \mid \bld m^{(a)} \overset{ind.}\sim \mbox{Unif}(\m Z_{N_a,K, \blds m^{(a)}}), \quad \bld m^{(a)} = \mbox{rd}\big(N^{(a)}, \bld w^{(a)}\big), \quad a \in [r],    
\end{align} 
where we first set the cluster occupancy vector $\bld m^{(a)} = \mbox{rd}\big(N^{(a)}, \bld w^{(a)}\big)$ for each $a\in[r]$, and then draw $\bld z^{(a)}$ uniformly from $\m Z_{N_a,K, \blds m^{(a)}}$. This layer in our model is critically different from a usual specification in a hierarchical Dirichlet process prior \citep{doi:10.1198/016214506000000302}, and is necessary in our endeavour to embed the idea of balance in the resulting clusters. We call the induced prior on $\bld m^{(a)}$ implied by the hierarchy $\bld m^{(a)} = \mbox{rd}\big(N^{(a)}, \bld w^{(a)}\big), w^{(a)} \sim \mbox{Dir}(\alpha_0 \, \bld \beta)$ a {\em lifted Dirichlet} prior with parameters $N^{(a)}, \alpha_0, \beta$. Compared to the more standard Dirichlet-Multinomial prior, where $\bld m^{(a)}$ is additionally sampled from a Multinomial distribution with total count $N^{(a)}$ and probability vector $\bld w^{(a)} \sim \mbox{Dir}(\alpha_0 \, \bld \beta)$, the randomness in a lifted Dirichlet prior is entirely controlled by $\bld w^{(a)}$, enabling tighter control on the cluster sizes across $a$. For large $N^{(a)}$, a lifted Dirichlet distribution provides a close approximation to a Dirichlet-Multinomial due to concentration of measure of Multinomials \citep{qian2020concentration}. In Figure \ref{plot:KL_Y1Y2}, we offer a visual illustration of this phenomenon in the $K=2$ case, where we set $\gamma_k = \alpha_0 \beta_k$ for $k =1, 2$, and plot a heatmap of symmetrized KL distances between lifted Beta and Beta-Binomial distributions against $(\gamma_1, \gamma_2)$ for four different choices of $N$. Clearly, as $N$ increases, the approximation uniformly improves across values of the Beta hyperparameters $\gamma_1$ and $\gamma_2$. 

The final part of our hierarchical model constitutes specifying component-wise distributions:
\begin{equation}\label{hdpmm_3}
\begin{split}
 & \bld x_i^{(a)} \mid z_i^{(a)} = k, \bld \phi_k^{(a)} \stackrel{\textit{ind.}}{\sim} f_{obs}(\cdot \mid \bld \phi_k^{(a)}), \quad i \in [N_a], \ a \in [r]\\
& \bld \phi_k^{(a)} \mid \bld \phi^{(a)} \stackrel{\textit{ind.}}{\sim} f_{pop}(\cdot \mid \bld \phi^{(a)}), \quad k \in [K], \ a \in [r],  \quad
\bld \phi^{(a)} \stackrel{\textit{i.i.d}}{\sim} f_{atom}, \quad a \in [r],
\end{split}
\end{equation}
where $\bld \phi_k^{(a)}$ are parameters specific to $k$-th cluster and $a$-th attribute, which are drawn independently from a common population. The equations \eqref{hdpmm_1} -- \eqref{hdpmm_3} complete the specification of our hierarchical model, which we call a {\em Hierarchical Fair-Dirichlet process} (HFDP). For concreteness, we focus on a Gaussian generative model with a conjugate NIW prior on the mean-covariance in \eqref{hdpmm_3}, i.e., set $f_{obs}(\cdot \mid \bld \phi_k^{(a)}) = \mbox{N}_d\big(\bld\mu^{(a)}_k,\ \Sigma^{(a)}_k\big)$ where $\phi_k^{(a)} = (\bld\mu^{(a)}_k,\ \Sigma^{(a)}_k)$; and  $f_{pop}(\cdot \mid \bld \phi^{(a)}) = \mbox{NIW}(\bld\mu^{(a)}_0, \ \lambda^{(a)}_0,\ \Lambda^{(a)}_0,\ \nu^{(a)}_0)$, where $\phi^{(a)} = \bld\mu^{(a)}_0, \ \lambda^{(a)}_0,\ \Lambda^{(a)}_0,\ \nu^{(a)}_0)$ 
with $\bld\mu^{\star}, \ \Lambda^{\star}$ suitably fixed. Finally, we collect $\bld\mu^{(a)} = \{\bld\mu^{(a)}_1,\ldots,\bld\mu^{(a)}_K\} $ and $\Sigma^{(a)} = \{\Sigma^{(a)}_1,\ldots,\Sigma^{(a)}_K\}$ where $a\in[r]$. Under the above model and prior specification, our goal is to learn the clustering indices $\{\bld z^{(a)}, a\in[r]\}$ respecting the notion of approximate balance, and avoid the need for statistically inefficient posterior sample sanitation, such as the one described in  text leading up to Theorem \ref{th1} in section \ref{ssec:principles}. Extensions to non-Gaussian likelihoods are straightforward. 

\section{Posterior Computation}\label{ssec:blocked_gibbs}
Sampling from the the posterior obtained from \eqref{hdpmm_1}-\eqref{hdpmm_3} encounters challenges due to (a) the lack of conjugacy of the shared weight vector $\beta$ and $\alpha_0$ and (b) sampling from the cluster occupancy vectors $\bld m^{(a)}$ in large discrete spaces with combinatorial complexity.   
The bottleneck (a) is alleviated by noting that the Gamma prior on $\alpha_0$ allows us to break the dependence of the shared mixing proportions $\beta$. This permits independent updates of certain log-concave random variables in a block, adapting a recently developed blocked Gibbs sampler for the hierarchical Dirichlet processes \citep{das2023blocked}. 
The second bottleneck is carefully circumnavigated via a novel scheme for non-uniform sampling from the space of binary matrices with fixed margins, utilizing techniques from integer-valued optimal transport towards constructing proposals. Starting with the joint distribution of all parameters described in equations \eqref{hdpmm_1} -- \eqref{hdpmm_3}, we analytically integrate out cluster specific parameters $\{\bld \phi_k,\ k\in[K] \}$ and shared concentration parameter $\alpha_0$, wherever possible, to ensure improved mixing. The details of the derivation of the sampler are deferred to the \textcolor{purple}{supplementary materials}, and the key computational bottlenecks and features of the  algorithm are presented in sequel.

Letting $\theta\mid -$ denote the full conditional distribution of a parameter $\theta$ given other parameters and the data, a collapsed Gibbs sampler (or  an MC-EM algorithm), cycles through the following steps. First, we sample from $[\alpha_0\mid - ]$ via Metropolis-within-Gibbs, sample from $\bld\beta\mid -$ adapting \citep{das2023blocked} to our present setup, and then sample $[\bld w^{(a)}\mid -]$ from independent Dirichlet distributions.
The marginal conditional of clustering indices $[\bld z^{(a)}\mid -], \ a\in[r]$, integrating out population parameters, is
\begin{equation}\label{eq:non_u}
    [\bld z^{(a)}\mid -] \ \propto\ \prod_{k=1}^K \frac{\Gamma_{d}(\nu^{(a)}_{k}/2)\ (\lambda^{(a)}_{0})^{d/2}\ |\Lambda^{(a)}_{0}|^{\nu^{(a)}_{0}/2}}{\Gamma_{d}(\nu^{(a)}_{0}/2)\ (\lambda^{(a)}_{k})^{d/2}\ |\Lambda^{(a)}_{k}|^{\nu^{(a)}_{k}/2}}, \quad \bld z^{(a)} \in \m Z_{N_a,K, \blds m^{(a)}} \quad a\in[r].
\end{equation}
Sampling \eqref{eq:non_u} poses a difficult combinatorial problem and is the most substantial computational bottleneck in our algorithm. We circumnavigate this issue via recasting the problem as a non-uniform sampling task from the space of binary matrices with fixed margin \citep{miller_exact, BRUALDI198033, curveball}, and propose a novel sampling scheme equipped with an MH proposal based on integer-valued optimal transport \citep{Villani2008OptimalTO}.

\subsection*{Optimization}
As an intermediate step, we first develop a computationally convenient MC-EM algorithm \citep{10.2307/1391097}, where instead of sampling from $[\bld z^{(a)}\mid -], \ a\in[r]$, we update the chain with the posterior mode of $[\bld z^{(a)}\mid -], \ a\in[r]$. In particular,  to compute the posterior mode of $[\bld z^{(a)}\mid -], \ a\in[r]$, we go over the following steps.  
\textbf{(i)} First, for $a\in[r]$, we calculate the current component specific means and variances $\bld \mu^{(a)}_{k, \star}, \Sigma^{(a)}_{k, \star}$ for all $a\in[r], k\in[K]$.
\textbf{(ii)} Next,  we  define the $N_a\times K$  cost matrix $\mbox{L}^{(a)}= ((l_{ik})) = \big(\big(-\log \mbox{N}_d(\bld x^{(a)}_{i}\mid \bld\mu^{(a)}_{k,\star}, \Sigma^{(a)}_{k, \star}))$, column sum vectors $\bld c^{(a)}=\bld m^{(a)}$ and row sum  vectors $\bld r^{(a)} = \bld 1_{N_a}$, where $\bld 1_{N_a}$ is a vector of $N_a$ $1$s.
\textbf{(iii)} Next, given the two vectors $\bld r^{(a)}, \bld c^{(a)}$, we define the polytope of $K\times N_a$ binary cluster membership matrices $\mbox{U}(\bld r^{(a)}, \bld c^{(a)}):=\{\mbox{B}\mid \mbox{B}\bld 1_{N_a} = \bld r^{(a)};\ \mbox{B}^{\T}1_{K} = \bld c^{(a)}\}$, and solve the constrained binary optimal transport problem \citep{Villani2008OptimalTO}
$\mbox{B}^{(a)}= \mbox{\rm argmin}_{\mbox{B}\in \mbox{U}(\bld r^{(a)},\ \bld c^{(a)})} \langle\mbox{B}, \mbox{L}^{(a)} \rangle$ where $\langle\mbox{B}, \mbox{L}^{(a)}\rangle = \rm{tr}(\mbox{B}^\T \mbox{L}^{(a)})$. Finally, we obtain the modal clustering indices $\bld z^{(a)}$ from the binary cluster membership matrices $\mbox{B}^{(a)}, a\in[r]$, completing the MC-EM algorithm. 

\subsection*{Non-uniform Sampling of Fixed Margin Binary Matrices}
We now complete the details of our complete Gibbs sampler which replaces the previous optimization with sampling. Uniform distribution on binary matrices with fixed margins arises in many applications in  neurophysiology, sociology, psychometrics and ecology. Both exact \citep{miller_exact} and approximate uniform sampling methods from the space of binary matrices  with fixed margins are now well-established. The approximate sampling algorithms are based on checker board swap \citep{BRUALDI198033}, curve-ball \citep{curveball} or rectangular loop moves \citep{recloop}. However, \eqref{eq:non_u} requires sampling according to a non-uniform probability distribution defined by a weight matrix. To that end, we adapt \citep{recloop} to introduce rectangular loop moves in the non-uniform case to improve mixing. We explain this adaptation below. 

Given fixed row sums $\mathbf{r} = (r_1,\ldots, r_u)^{\T}$ and column sums $\mathbf{c} = (c_1,\ldots, c_v)^{\T}$, we denote the collection of all $u\times v $ binary matrices by $\mbox{U}(\mathbf{r}, \mathbf{c})$. Further, we denote by $\Omega=(\omega_{ij})\in [0,\infty)$ a  non negative weight matrix representing the relative probability of observing a count of $1$ at the $(i,j)$-th cell. Then the likelihood associated with the observed binary matrix $H\in\mbox{U}(\mathbf{r}, \mathbf{c})$ as
$\mbox{P}(H) = (1/\kappa) \prod_{i,j} \omega_{ij}^{h_{ij}},\ \kappa = \sum_{H\in\mbox{U}(\mathbf{r}, \mathbf{c})}\ \prod_{i,j} \omega_{ij}^{h_{ij}}$. 
Let  $\mbox{U}^{\prime}(\mathbf{r}, \mathbf{c}) = \{H\in\mbox{U}(\mathbf{r}, \mathbf{c}):\ P(H)>0\}$ denote the subset of matrices in $\mbox{U}(\mathbf{r}, \mathbf{c})$ with positive probability. Then, for  $H_1,H_2\in\mbox{U}^{\prime}(\mathbf{r}, \mathbf{c})$,  the relative probability of the two observed matrices is
$\mbox{P}(H_1)/\mbox{P}(H_2) = \prod_{\{i,j: h_{1, ij}=1, h_{2, ij}=0\}} \ \omega_{ij}^{h_{1, ij}}/ \prod_{\{i,j: h_{1, ij}=0, h_{2, ij}=1\}} \   \omega_{ij}^{h_{2, ij}}$.
With these notations, we are all set to introduce a \emph{Weighted Rectangular Loop Algorithm} (W-RLA)  for non-uniform sampling from the space of fixed margin binary matrices $H\in\mbox{U}(\mathbf{r}, \mathbf{c})$, given the weight matrix $\Omega=(\omega_{ij})\in [0,\infty)$.  To that end, let us first record that the identity matrix of order $2$, and the $2\times2$ matrix  with all zero  diagonal entries and all one off-diagonal entries, are referred to as \emph{checker-board} matrices. 

W-RLA is described as follows. \textbf{(1)} At iteration $t=0$, provide an initial binary matrix $A_0$ that respects the margin constraints, and total number of iterations $T$. \textbf{(2)} Increment $t\to t+1$, and \emph{(a)} choose one row and one column $(r_1,c_1)$ uniformly at random. \emph{(b)} \emph{If} $A_{t-1}(r_1,c_1) = 1$, choose a column $c_2$ at random among all the $0$ entries in $r_1$, and a row $r_2$ at random among all the $1$ entries in $c_2$. \emph{Else} choose a row $r_2$ at random among all the 1 entries in $c_1$, and a column $c_2$ at random among all the $0$ entries in $r_2$. \textbf{(3.1)} \emph{If} the sub-matrix extracted from $r_1, r_2, c_1, c_2$ is a checkerboard unit -- obtain $B_{t}$ from $A_{t-1}$ by swapping the checker-board;  calculate $p_{t}=\mbox{P}(B_{t})/[\mbox{P}(B_{t}) + \mbox{P}(A_{t-1})]$; draw $r_t\sim \mbox{Bernoulli}(p_{t})$; and  \emph{If} $r_t = 1$, then set $A_{t}=B_{t}$; \emph{else} set $A_{t}=A_{t-1}$.
\textbf{(3.2)} \emph{If} the sub-matrix extracted from $r_1, r_2, c_1, c_2$ is \emph{not} a checkerboard unit, set $A_{t}=A_{t-1}$. This completes the description of W-RLA. Theorem \ref{th:validity_weighted_recloop} ensures  the validity of the scheme presented in  the W-RLA, i.e, it demonstrates that the Markov chain converges to the correct stationary distribution as the number of iterations $T\to\infty$. The proof the Theorem \ref{th:validity_weighted_recloop} is deferred to the \textcolor{purple}{supplementary material}.
\vspace{-2mm}
\begin{theorem}\label{th:validity_weighted_recloop}
Given an weight matrix $\Omega$ with $\omega_{ij} > 0$, and fixed row-sum and column sum vectors $(\bld r, \bld c)$,  the Weighted Rectangular Loop Algorithm (W-RLA) generates a
Markov chain with stationary distribution given by 
\begin{align*}
  \mbox{P}(H) = (1/\kappa) \prod_{i,j} \omega_{ij}^{h_{ij}},\ \kappa = \sum_{H\in\mbox{U}(\mathbf{r}, \mathbf{c})}\ \prod_{i,j} \omega_{ij}^{h_{ij}}.  
\end{align*}
\end{theorem}
In our blocked Gibbs sampler, the final step involves sampling binary cluster membership matrices, given an weight matrix $\Omega = \mbox{L}^{(a)}= ((l_{ik})) = \big(\big(-\log \mbox{N}_d(\bld x^{(a)}_{i}\mid \bld\mu^{(a)}_{k,\star}, \Sigma^{(a)}_{k, \star}))$, column sum vector $\bld c^{(a)}=\bld m^{(a)}$, row sum vector $\bld r^{(a)} = \bld 1_{N_a}$, that are all fixed conditional on the remaining model parameters. In order to sample from the full conditional of $[\bld z^{(a)}\mid -], a\in [r]$, first we obtain a proposal for cluster membership matrix $\mbox{B}^{(a)}_{\rm prop}$ by solving the constrained binary optimal transport \citep{Villani2008OptimalTO} problem 
$\mbox{B}^{(a)}_{\rm prop}= \mbox{\rm argmin}_{\mbox{B}\in \mbox{U}(\bld r^{(a)},\ \bld c^{(a)})} \langle \mbox{B}, \mbox{L}^{(a)}\rangle$ as in the optimization scheme, and then mutate the proposal via the rectangular loop scheme to get $\mbox{B}^{(a)}_{\rm rec-loop}$. Finally, we set the updated $\bld z^{(a)}, a\in[r]$ to cluster indicators corresponding to $\mbox{B}^{(a)}_{\rm rec-loop}$ with probability $\mbox{P}({B}^{(a)}_{\rm rec-loop})/[\mbox{P}({B}^{(a)}_{\rm rec-loop}) + \mbox{P}(\mbox{B}^{(a)}_{\rm prop})]$ and to cluster indicators corresponding to $\mbox{B}^{(a)}_{\rm prop}$ other-wise.

\section{Experiments}

\begin{figure}[!htbp]
\begin{center}
\includegraphics[width=13cm, height =3cm]{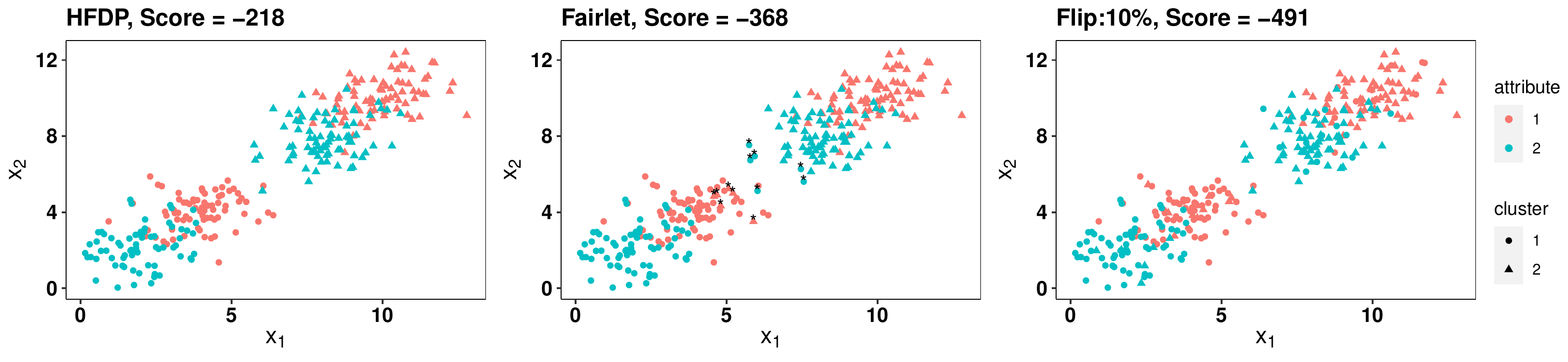} 
\centering{{\caption{\emph{For one instance in the simulation set up (A.1), the left and middle panel present clustering configuration via HFDP and  fair k-means\citep{NIPS2017_978fce5b}, respectively. The differences between the two configurations are marked with black "*". Non-optimal exchange of points in fair k-means leads to substantial decrease in fair-score. The right panel presents the clustering configurations obtained via randomly flipping $10\%$ of the MAP of the HFDP clustering indices, to demonstrate that fair-score is low for non-sensical clustering configurations.}}\label{fig:fair_score_validity}}}.
\end{center}
\end{figure}
\subsection{Simulation study 1 (Two/Multi-color Case)}
We generate data from the model in \eqref{hdpmm_1}-\eqref{hdpmm_3} with fixed $(b, g)$ that yields a small $\alpha_0$, which in turn yields low balance in the clusters. We fix the true number of clusters $K_{true}=2$, and attribute specific sample sizes $N^{(1)}=N^{(2)}=200$. We consider the cases

\begin{figure}
\begin{center}
\includegraphics[width=14cm, height =2.0cm]{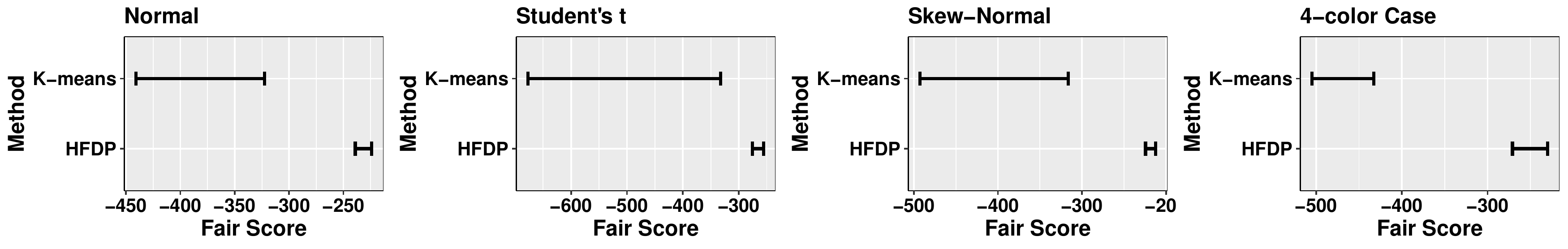} 
\centering{{\caption{\emph{Two/Multi-color case. The MAP of the HFDP fares better compared to  fair-clustering with fairlets \citep{NIPS2017_978fce5b} (termed as  K-Means) in two color case, and  the method in \citet{https://doi.org/10.48550/arxiv.2002.07892} (termed as  K-Means) in multi-color case, in terms of the fair-score \eqref{criterion}; and automatically estimates the number of clusters.}}\label{fig:two_multi_color}}}.
\end{center}
\end{figure}

\begin{itemize}
    \item \textbf{(A.1) Well-specified case (multivariate normal components).} In the first cluster, the individuals with $a=1$ are generated from $N_2(\mu_{11}, S)$ and individuals with $a=2$ are generated from $N_2(\mu_{21}, S)$.  In the second cluster, individuals with $a=1$ are generated from $N_2(\mu_{12}, S)$ and  individuals with $a=2$ are generated from $N_2(\mu_{22}, S)$, where $\mu_{11}=(4,4)^{\prime},\  \mu_{21} = (2,2)^{\prime},\ \mu_{12} = (10,10)^{\prime},\ 
 \mu_{22} = (8,8)^{\prime}$, and $S = 3\times[\rho 11^{\T} + (1-\rho)I_2]$ with $\rho = 0.3$.

 \item \textbf{(A.2) Mis-specified case 1 (multivariate t components).} We follow the same scheme as before, except for simulating from multivariate $t$-distributions with centers $(\mu_{11},\mu_{21},\mu_{12}, \mu_{22})$, scale $S$, and degrees of freedom $4$.

 \item \textbf{(A.3) Miss-specified case 2 (multivariate skew normal components).}
We follow the same scheme as before, except for simulating from multivariate skew normal distributions \citep{10.1093/biomet/83.4.715} with centers $(\mu_{11},\mu_{21},\mu_{12}, \mu_{22})$, scale $S$, and the skewness parameter $\alpha = (1, 1)^{\T}$.  In Figure \ref{fig:two_multi_color},  fair-clustering with fairlets \citep{NIPS2017_978fce5b} (termed as  K-Means) is compared with HFDP via repeated simulations both in well-specified and mis-specified set ups. Across all the numerical experiments, the MAP of the HFDP fares substantially better in terms of the fair-score \eqref{criterion}. Further HFDP estimates the number of clusters automatically, but the number of clusters  for the methods in \citep{NIPS2017_978fce5b, https://doi.org/10.48550/arxiv.2002.07892} are set deterministically  at the value estimated by the MAP of the HFDP. 

\item \textbf{B. Multiple colors.} 
We generate data from the model in \eqref{hdpmm_1}-\eqref{hdpmm_3} with number of colors $r=4$, sample sizes $N^{(1)}=N^{(2)}=N^{(3)} = N^{(4)} =200$,  true number of clusters $K_{true}=2$, and large $b$.  In the first cluster, the individuals with $a=1$ are generated from $N_2((4,4)^{\prime}, S)$, the individuals with $a=2$ are generated from $N_2((2,2)^{\prime}, S)$, the individuals with $a=3$ are generated from $N_2((0,0)^{\prime}, S)$, the individuals and with $a=4$ are generated from $N_2((-2,-2)^{\prime}, S)$.  In the second cluster, individuals with $a=1$ are generated from $N_2((10,10)^{\prime}, S)$, individuals with $a=2$ are generated from $N_2((8,8)^{\prime}, S)$, individuals with $a=2$ are generated from $N_2((6, 6)^{\prime}, S)$, and individuals with $a=4$ are generated from $N_2((4, 4)^{\prime}, S)$ where $S = 3\times[\rho 11^{\T} + (1-\rho)I_2]$ with $\rho = 0.3$.  Finally, we carry out the clustering via the algorithm in section \ref{ssec:blocked_gibbs}, fixing the values of the hyper-parameters $g, \mu_{p0}, \lambda_{p0}, \nu_{p0}, \Lambda_{p0}$ such that these parameters do not impact our analysis.  In Figure \ref{fig:two_multi_color},  the method in \citet{https://doi.org/10.48550/arxiv.2002.07892} (termed as  K-Means) is compared with HFDP via repeated simulations. Across all the numerical experiments, the MAP of the HFDP fares substantially better in terms of the fair-score \eqref{criterion}, as well as automatically estimates the number of clusters.
\end{itemize}

\subsection{ Simulation study 2 (Imperfect Knowledge of Protected Attribute)}\label{Simulation_Set4}
\begin{figure}
\begin{center}
\includegraphics[width=12cm, height =2.0cm]{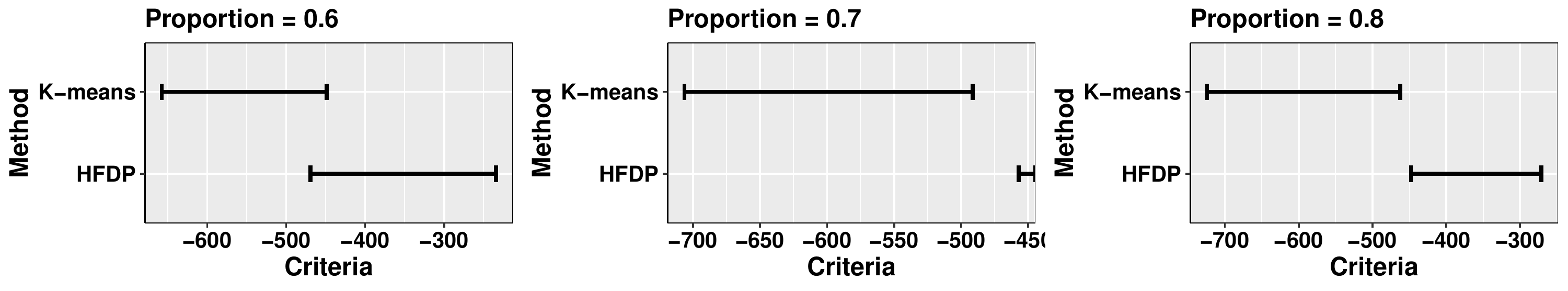} 
\centering{{\caption{\emph{Imperfect knowledge of protected attribute.  The MAP of the HFDP fares better  compared to probabilistic fair clustering \citep{NEURIPS2020_95f2b84d} (termed as  K-means), in terms of the fair-score \eqref{criterion} in repeated simulations with varying $p_{acc}$.}}\label{fig:imperfect}}}.
\end{center}
\end{figure}

Much of the prior work on fair clustering assumes complete knowledge of group membership.  \citet{NEURIPS2020_95f2b84d}  assumed imperfect knowledge of group membership through probabilistic assignments, i.e, for individual $i\in[N]$, the protected label $a_{i} = a$ with probability $p^{(a)}_{i}$ (known) and $\sum_{a=1}^r p^{(a)}_{i} = 1, \ i\in[N]$. The goal is to balance the \emph{expected color} in each cluster. Our fully probabilistic set up enables us seamlessly achieve this via augmenting every cycle of our computational scheme by an additional sampling step as follows $a_i = a\quad \text{with probability}\ p^{(a)}_{i},\ a\in[r],\ i\in[N]$. To study numerical efficacy,  we  first generate data exactly as in \emph{simulation 1, case A.1}; then retain the label of $A$ with probability $p_{acc}\in\{0.7, 0.8, 0.9\}$ for each of the individuals, and swap with probability $1-p_{acc}$. In Figure \ref{fig:imperfect}, we compare HFDP with probabilistic fair clustering \citep{NEURIPS2020_95f2b84d} (termed as  K-means) via repeated simulations with varying $p_{acc}\in\{0.7, 0.8, 0.9\}$. Across all the numerical experiments, the MAP of the HFDP fares substantially better in terms of the fair-score \eqref{criterion}.

\subsection{\bf Benchmark Datasets}
We compare the performance of HFDP relative to the existing methodologies on popular bench mark data sets from the UCI repository \citep{Dua:2019}, considered before in the literature \citep{NIPS2017_978fce5b, https://doi.org/10.48550/arxiv.2002.07892, NEURIPS2020_95f2b84d}: 
\begin{itemize}
    \item \textbf{(1) Diabetes Data.}
We chose numeric attributes such as age, time in hospital, to represent points in the euclidean space and gender as the sensitive dimension. We sub-sampled $1000$ individuals from the data-set, with proportion of two genders equal to $(0.474, 0.526)$, i.e the target balance is $0.90$. HFDP estimates the modal number of clusters to be $5$, corresponding balance $0.898$ and fair-score \eqref{criterion}  equal to $-114.74$. Fair clustering with fairlets with fixed $K=5$ yields value of the fair-score \eqref{criterion} equal to $-651.84$.

\item \textbf{(2) Portuguese Banking Data.}
We chose numeric attributes such as age, balance, and duration to represent points in the
Euclidean space, we aim to cluster to balance married and not married clients. We sub-sampled the data set
to $1000$ records, with proportion married and not married clients equal to $(0.626, 0.374)$, i.e the target balance is $0.60$. HFDP estimates the modal number of clusters to be $5$, corresponding balance $0.593$ and the fair-score \eqref{criterion}  equal to $-176.85$. Fair clustering with fairlets with fixed $K=5$ yields the fair-score \eqref{criterion} equal to $-639.12$.

\item \textbf{(3) Credit Card Data.} 
We chose numeric attributes such as age, credit limit, to represent points in the euclidean space and marital status (married, unmarried, others) as the sensitive dimension. We sub-sampled $600$ individuals from the data-set, such that the target balance is $0.99$. HFDP estimates the modal number of clusters to be $3$, corresponding balance $0.99$ and value of the fair-score \eqref{criterion}  $-86.19$. The algorithm in \citet{https://doi.org/10.48550/arxiv.2002.07892} with fixed $K=3$ yields value of the fair-score \eqref{criterion}  equal to $-271.09$.
\end{itemize}

\bibliographystyle{apalike}
\bibliography{fairness}

\clearpage
\beginsupplement
\begin{center}
   \section*{Supplementary material to \\ ``Fair Clustering via Hiearchical Fair Dirichlet Process"} 
  Abhisek Chakraborty, Anirban Bhattacharya, Debdeep Pati\\
 Department of Statistics, Texas A\&M University, College Station, TX, USA
\end{center}

\beginsupplement
Section \ref{proof_th1} contains the detailed proof of Theorem 1, presented in section 2.2 in the main document. Sub-section \ref{sup_posterior_comp} contains the detailed derivation of the posterior computation steps, presented in section 4 in the main document. Sub-section \ref{sup_wrla} records  the Weighted Rectangular Loop algorithm in complete details, presented in section 4 in the main document. Section \ref{extensions} discusses straight forward extension of our model based procedure to accommodate practical complexities -  mixed data type, censoring, missing values, etc. Section \ref{performance} provides additional numerical results to demonstrated the validity of the fair-score, introduced in equation (3) in sub-section (2.1) in the main document, in assessing the clustering performance under balance constraints.

\section{Proof of Theorem 1 section 2.2 in the main document}\label{proof_th1}
\begin{proof}
The proof proceeds via three key steps. \textbf{Step 1} deals with sample quantities in the left hand side of the statement, and \textbf{Step 2} simplifies the population quantities in the right hand side of the statement. Finally, \textbf{Step 3} presents the limiting arguments to prove the equivalence result.

\subsection*{Step 1 (sample quantities)}
We observe data $(\bld X, \bld A)$, and want to impute the clustering indices  $\bld Z \in[K]^N$.  Suppose
$\mathcal{P}_{\blds Z\mid \blds X,\blds A}(\bld z \mid \bld x,\bld a)\ \propto\ \mathbf{m}_N(\bld z)$
is the conditional  posterior distribution of $\bld Z\mid \bld X, \bld A$. Under common hierarchical model specifications, e.g \citet{doi:10.1198/016214506000000302}, the marginal posterior of clustering configuration $\mathbf{m}_N(\bld z)$ often does not factorise over $k\in[K]$, owing to marginalization with respect to lower level parameters. For the sake of this proof, we assume that the lower level parameters are set at reasonable values, say obtained via an empirical Bayesian procedure. Then, we have 
$\mathcal{P}_{\blds Z\mid \blds X,\blds A}(\bld z \mid \bld x,\bld a)\ \propto\ \mathbf{m}_N(\bld z) = \prod_{a=1}^r\prod_{k=1}^K \pi_{a, k}^{N_{a, k}}$, where $N_{a, k} = \sum_{a=1}^r\sum_{k=1}^K \mathbf{1}(z_i=k, a_i = a)$ for all $a\in[r],\ k\in[K]$. Note that, $((\pi_{a, k}))$  encode the information in $(\bld x, \bld a)$.

We fix $\varepsilon\geq 0$.  Assume that the restricted conditional distribution of $(\bld X,\bld A, \bld Z)$ is denoted by $\mathcal{P}^{\prime}_{\blds Z\mid \blds X,\blds A}$, takes the form:
\begin{align}
    \mathcal{P}^{\prime}_{\blds Z\mid \blds X, \blds A}(\bld z\mid \bld x, \bld a)
    \ \propto\
    \frac{\mathbf{m}_N(\bld z)}{W^{(N)}}\times
    \mathbf{1}[\bld z^{(N)}\in \mathbb{Z}^{(fair)}(\varepsilon)],
\end{align}
where
$W^{(N)} = \sum_{\bld z^{(N)}\in \mathbb{Z}^{(fair)}(\varepsilon)} \mathbf{m}_N(\bld z)$ . We can sample from $\mathcal{P}_{\blds Z\mid \blds X,\blds A}$ (say, MCMC samples), and retain the sample if it satisfies $\bld Z\in\mathbb{Z}^{(fair)}(\varepsilon)$. For the retained sample, we calculate $\pi^{(N)}_{a, k}(\varepsilon) =(N_{a, k}/N)$ and collect in  a matrix $\bld \pi^{(N)}(\varepsilon) = ((\pi^{(N)}_{a, k}(\varepsilon)))$.
Further, we note that the dual objective function for maximization of the restricted likelihood of $\mathcal{P}^{\prime}_{\blds Z\mid \blds X, \blds A}(\bld z\mid \bld x, \bld a)$  takes the form:
\begin{align}\label{eqn:dual_data}
 \mathcal{T}(\bld z) 
 &:=  (1/N)\log\mathbf{m}_N(\bld z) - \lambda\ \mbox{KL}(\mathcal{P}_{\blds A}\times\mathcal{P}_{\blds Z}\ ||\ \mathcal{P}_{\blds A, \blds Z}) \notag\\
 &= (1/N)\sum_{a=1}^r\sum_{k=1}^K N_{a, k}\log\pi_{a, k} - \lambda\ \mbox{KL}(\mathcal{P}_{\blds A}\times\mathcal{P}_{\blds Z}\ ||\ \mathcal{P}_{\blds A, \blds Z}),
\end{align}
where $\lambda\geq 0$ is the Lagrange multiplier that depends on $\varepsilon$. 

\subsection*{Step 2. (population quantities)}
Next,  we uncover the quantities in Definition of pseudo true fair clustering distribution $\mathcal{P}^{\varepsilon}_{Z_{\rm PT-FCD}}$ in sub-section (2.1) in the main document. For the true generative model $\mathcal{P}^{\star}_{A, Z, X}$, we denote  $T = Z\mid X, A=a\sim \mbox{Categorical}(\pi^{\star}_{a,1},\ldots,\pi^{\star}_{a, K})$, and collect $\bld \pi^{\star} = ((\pi^{\star}_{a, k}))$. For any $\mathcal{P}_{A, Z, X}\in \mathbb{P}^{(1)}_{\varepsilon}$, we denote $V_{\varepsilon} = Z\mid X, A = a \sim \mbox{Categorical}(\Tilde{\pi}_{a, 1}(\varepsilon),\ldots,\Tilde{\pi}_{a, K}(\varepsilon))$ and collect $\Tilde{\bld \pi}(\varepsilon) = ((\Tilde{\pi}_{a, k}(\varepsilon)))$. To minimize
$\mbox{\rm KL}(\mathcal{P}^{\star}_{Z\mid X, A}\ ||\ \mathcal{P}_{Z\mid X, A}) =  \mathbb{E}_{{P}^{\star}}(\log\mathcal{P}^{\star}_{Z\mid X, A}) - \mathbb{E}_{{P}^{\star}}(\log\mathcal{P}_{Z\mid X, A})$ with respect to  $\mathcal{P}_{A, Z, X}\in \mathbb{P}^{(1)}_{\varepsilon}$, we can equivalently maximize 
\begin{align}
   \sum_{k=1}^K \pi^{\star}_{a, k} \log\Tilde{\pi}_{a, k}(\varepsilon) 
    - \lambda\ \mbox{KL}(\mathcal{P}_{ A}\times\mathcal{P}_{ Z}\ ||\ \mathcal{P}_{ A, Z}), 
\end{align}
with respect to $\Tilde{\bld \pi}(\varepsilon)$. 

\subsection*{Step 3 (equivalence)}
First, we uncover the quantities in Definition  of maximum-a-posteriori fair cluster ${\bld z}^{\varepsilon}_{\rm MAP-FC}$ in sub-section (2.2) in the main document. We observe that,
\begin{equation}\label{th1_eq1}
\begin{split}
  \pi^{(N)}_{a\mid k} 
     &=\frac{\sum_{i=1}^N \mathbf{1}(z_i = k, a_i = a)}{\sum_{i=1}^N \mathbf{1}(z_i = k)}
    =\frac{[\sum_{i=1}^N \mathbf{1}(z_i = k, a_i = a)]/N}{[\sum_{i=1}^N \mathbf{1}(z_i = k)]/N}\notag\\
    &\xrightarrow{a.s}\frac{\mathcal{P}_{Z, A}(Z= k, A = a)}{\mathcal{P}_{Z}(Z = k)} = \mathcal{P}_{ A\mid Z}(A=a\mid Z=k),  
\end{split}
\end{equation}
$\forall \ a\in\mathcal{A},\ \forall\ k$ as $N\to\infty$, by the \emph{Strong Law of Large Numbers}.  Then, by Equation \eqref{th1_eq1},  there exists a $\mathcal{P}_{X,Z,A}\in\mathbb{P}^{(1)}_{\varepsilon}$ such that
$\mathcal{P}^{\prime}_{\blds X, \blds Z, \blds A}\xrightarrow{a.s} \mathcal{P}_{X,Z,A}$ as $N\to\infty$. Recall that we defined,
$
    \mathbb{Z}^{(fair)}(\varepsilon) = \big\{\bld Z\in \mathbb{Z}:\mbox{KL}(\mathcal{P}_{\blds A}\times\mathcal{P}_{\blds Z}\ ||\ \mathcal{P}_{\blds A, \blds Z})\leq \varepsilon\big\}.
$
Then for any $\bld Z \in \mathbb{Z}^{(fair)}(\varepsilon)$, we have 
\begin{align}
&\mbox{KL}(\mathcal{P}_{\blds A}\times\mathcal{P}_{\blds Z}\ ||\ \mathcal{P}_{\blds A, \blds Z})\leq \varepsilon
\equiv
\sum_{a, z} \pi^{(N)}_{a}\pi_{z}^{(N)}\ \log\bigg[\frac{\pi^{(N)}_{a}\pi_{z}^{(N)}}{\pi^{(N)}_{a, z}}\bigg]\leq \varepsilon\ \forall\ N,
\end{align}
which by the \emph{Strong Law of Large Numbers} yields
$\mbox{KL}(\mathcal{P}_{ A}\times\mathcal{P}_{ Z}\ ||\ \mathcal{P}_{ A,  Z})\leq \varepsilon$ as $N\to\infty$, ($\star$).

Next, we note that under clustering consistency conditions, the unconstrained log marginal likelihood of $\bld z$, $(1/N)\log\mathbf{m}_N(\bld z)\xrightarrow{a.s} \sum_{k=1}^K {\pi}^{\star}_{a, k} \log\Tilde{\pi}_{a, k}(\varepsilon)$. This together with  $\star$ implies 
\begin{align}\label{eqn:dual_pop}
(1/N)\log\mathbf{m}_N(\bld z) - \lambda\ \mbox{KL}(\mathcal{P}_{\blds A}\times\mathcal{P}_{\blds Z}\ ||\ \mathcal{P}_{\blds A, \blds Z})\xrightarrow{a.s} 
    \sum_{k=1}^K {\pi}^{\star}_{a, k} \log\Tilde{\pi}_{a, k}(\varepsilon) - \lambda\ \mbox{KL}(\mathcal{P}_{ A}\times\mathcal{P}_{ Z}\ ||\ \mathcal{P}_{ A,  Z}),
\end{align}
as $N\to\infty$. 

The objective function $\mathcal{T}(\bld z)$ in Equation \eqref{eqn:dual_data} is maximised with respect to $\bld z$. The maxima is attained at a $\bld z$ with proportion of $(k,a)\in[K]\times[r]$ equal to $\bld\pi^{(N)}(\varepsilon)$, by definition. Suppose the objective function in Equation \eqref{eqn:dual_pop} is maximised with respect to $\Tilde{\bld \pi}(\varepsilon)$ at $\Tilde{\bld \pi}(\varepsilon) = \Tilde{\bld \pi}_m(\varepsilon)$.  Under suitable regularity conditions, we  invoke the \emph{Argmax Continuous Mapping Theorem} \citep{empprocess}, and we have $ \bld \pi^{(N)}(\varepsilon) \xrightarrow{P}\Tilde{\bld \pi}_m(\varepsilon) $. Recall that the map $\mathcal{R}:[K]^N\to[0, 1]^K$ takes a clustering configuration $\bld z\in[K]^N$ as input and outputs
the relative proportion of cluster indices $(1/N)[\sum_{i=1}^N \mathbf{1}(z_i = 1),\ldots, \sum_{i=1}^N \mathbf{1}(z_i = K)]^{\T}$. The marginal posterior of $(\bld Z \mid \bld X, \bld A)$ along with the reparameterization $\bld z\to \mathcal{R}(\bld z)$, yields the marginal posterior of $(\mathcal{R}(\bld Z)\mid \bld X, \bld A)$, denoted by $P_{\mathcal{R}(\blds Z) \mid \blds X, \blds A}$. Hence, we have the proof.    
\end{proof}


\section{Posterior Computation details}
\subsection{Full Conditionals in section 4 in the main document}\label{sup_posterior_comp}
For Dirichlet process mixture models, it is well known that blocked Gibbs sampling  has significant improvements in terms of both mixing and scalability over the Chinese restaurant process (CRP) based collapsed samplers. The blocked Gibbs algorithm replaces the infinite dimensional Dirichlet process prior by its finite dimensional approximation allowing the model
parameters to be expressed entirely in terms of a finite number of random variables, which are updated in blocks from simple multivariate distributions. Instead of adapting the CRF-based samplers to make them efficient and scalable, we adopt a blocked Gibbs sampler for our model in this paper, which will adhere to both improved mixing and scaling over large sample sizes. Starting with the joint distribution of all parameters described in equations (4)-(6) in the main document, along with specific choices, we intend to  analytically integrate out cluster specific parameters $\{\bld \phi_k,\ k\in[K] \}$, and intermediate level concentration parameter $\alpha_0$ , wherever possible,  to develop an efficient Collapsed Gibbs sampler, or an MC-EM algorithm, to sample from the posterior of the remaining parameters. The details of the derivation of the sampler are presented here.

The joint posterior of all parameters in equations (4)-(6) in the main document is as follows
\begin{align*}
&\pi(\bld\beta,\ \alpha_0,\{ \bld w^{(a)},\ \bld z^{(a)},\ \bld  \mu_{(a)},\ \Sigma_{(a)}\ \}_{a=1}^{r} \mid \{(\bld x_{i}, a_i)\}_{i=1}^N) \\
\propto\ &\    \bigg\{\prod_{a=1}^{r}\prod_{i=1}^{N_a} [\bld x_{(\sum_{j=1}^{a-1} n_j) + i}\mid \bld \mu^{(a)}, \Sigma^{(a)}, \bld z^{(a)}]\bigg\}\bigg\{\prod_{a=1}^r\prod_{k=1}^K [\bld\mu^{(a)}_k, \Sigma^{(a)}_k\mid \bld \mu^{(a)}_0, \Sigma^{(a)}_0]\bigg\}\\
& \ \bigg\{\prod_{a=1}^r [ \bld\mu^{(a)}_0, \Sigma^{(a)}_0]\bigg\}
\bigg\{\prod_{a=1}^r\bld [\bld z^{(a)}\mid \bld m^{(a)}]\bigg\} \ \bigg\{\prod_{a=1}^r [\bld w^{(a)}\mid \alpha_0,\bld \beta]\bigg\}\ [\alpha_0\mid g]\ [\bld\beta]\\
\propto &\ \bigg\{\prod_{a=1}^{r}\prod_{i=1}^{N_a} w^{(a)}_{z^{(a)}_i} \mbox{N}_d\big(\bld x_{(\sum_{j=1}^{a-1} n_j) + i}\mid\bld \mu^{(a)}_{z^{(a)}_i}, \Sigma^{(a)}_{z^{(a)}_i}\big)\bigg\} \\ 
& \ \bigg\{\prod_{a=1}^r\prod_{k=1}^K \mbox{NIW}(\bld\mu^{(a)}_k, \Sigma^{(a)}_k\mid \mu^{(a)}_0,  \lambda^{(a)}_0, \Lambda^{(a)}_0, \nu_{0}^{(a)})\bigg\} \bigg\{\prod_{a=1}^r\frac{\prod_{k=1}^K m^{(a)}_k !}{n_{a}!} \bigg\}\\
& \ \bigg\{\prod_{a=1}^r \mbox{Dir}(\bld w^{(a)}\mid \alpha_0 \bld\beta)\bigg\}\ \mbox{Gamma}(\alpha_0\mid g, b)\ \mbox{Dir}(\bld\beta\mid g/K,\ldots, g/K).
\end{align*}
In order to develop a computationally efficient blocked Gibbs sampler, we marginalise out the joint posterior with respect to the component specific parameters $\bld\mu^{(a)},\Sigma^{(a)}, a\in[r]$, and obtain the marginal posterior:
\begin{align*}
&\pi(\bld\beta,\ \alpha_0,\{ \bld w^{(a)},\ \bld z^{(a)}\}_{a=1}^{r} \mid \{(\bld x_{i}, a_i)\}_{i=1}^N) \\
\propto& \ \bigg\{\prod_{a=1}^r\prod_{k=1}^K \frac{\Gamma_{d}(\nu^{(a)}_{k}/2)\ (\lambda^{(a)}_{0})^{d/2}\ |\Lambda^{(a)}_{0}|^{\nu^{(a)}_{0}/2}}{\Gamma_{d}(\nu^{(a)}_{0}/2)\ (\lambda^{(a)}_{k})^{d/2}\ |\Lambda^{(a)}_{k}|^{\nu^{(a)}_{k}/2} }\bigg\}\\
& \ \bigg\{\frac{(\Gamma(\alpha_0))^r}{\prod_{k=1}^K(\Gamma(\alpha_0\beta_k))^r} \prod_{a=1}^r\prod_{k=1}^K (w_{k}^{(a)})^{\alpha_0 \beta_k + m^{(a)}_k - 1}\bigg\} \bigg\{\prod_{k=1}^K \beta_{k}^{\frac{g}{K}-1}\ \bigg\} \alpha_{0}^{g-1} e^{-b\alpha_0}\\
\end{align*}
where we have
\begin{align*}
& \lambda^{(a)}_{k} = \lambda^{(a)}_{0} + m^{(a)}_k ; \quad \nu^{(a)}_{k} = \nu^{(a)}_{0} + m^{(a)}_k;\\
& \bld\mu^{(a)}_k = \frac{\lambda^{(1)}_0 \bld\mu^{(a)}_0}{\lambda^{(a)}_{0} + m^{(a)}_k } + \frac{\sum_{i=1}^{N_a} \delta(z^{(a)}_i = k)\ \bld x_{(\sum_{j=1}^{a-1} n_j) + i}}{\lambda^{(a)}_{0} + m^{(a)}_k };\\
& \Lambda^{(a)}_{k} = \Lambda^{(a)}_{0} + \sum_{i=1}^{N_a} \delta(z^{(a)}_i =k)\ (\bld x_{(\sum_{j=1}^{a-1} n_j)+i} - \bar{\bld x}^{(a)}_{k})(\bld x_{(\sum_{j=1}^{a-1} n_j)+i} - \bar{\bld x}^{(a)}_{k})^\T + \\
&\frac{\lambda^{(a)}_{0} m^{(a)}_k}{\lambda^{(a)}_{0} + m^{(a)}_k }(\bar{\bld x}^{(a)}_{k} - \bar{\bld x})(\bar{\bld x}^{(a)}_{k} - \bar{\bld x})^{\T};\quad k\in [K],\  a\in[r]. \\
\end{align*}
The blocked parameter updates, and details of the computational strategies are enlisted next.
Letting $\theta\mid -$ denote the full conditional distribution of a parameter $\theta$ given other parameters and the data, the Gibbs sampler (or an \textbf{MC-EM} algorithm) cycles through the following steps, sampling parameters from their full conditional distributions:

\noindent{\bf Step 1:} To sample from $[\alpha_0\mid\cdot]$, we note that
\begin{align*}
    [\alpha_0\mid\cdot]\ 
    &\propto\ [\alpha_0\mid \bld\beta, \{\bld w^{(a)}\}_{a=1}^r] \\
    &\ \propto \bigg\{\frac{(\Gamma(\alpha_0))^r}{\prod_{k=1}^K(\Gamma(\alpha_0\beta_k))^r} \prod_{a=1}^r\prod_{k=1}^K (w_{k}^{(a)})^{\alpha_0 \beta_k}\bigg\} \alpha_{0}^{g-1} e^{-b\alpha_0}
\end{align*}
where $0<\alpha_0<\infty$. We can evaluate the density over a grid and sample from the resulting discrete distribution. In practice, our choice of the hyper-parameter $b$ dictates how far we need to go in the positive real line to construct our grid. Noteworthy, $\alpha_0$ is one-dimensional, and it can be easily sampled from its posterior via an Metropolis--Hastings update with independent proposals.

\noindent{\bf Step 2:} To sample from $[\bld\beta\mid\cdot]$, we note that 
\begin{align*}
    & [\bld\beta\mid\cdot]\ 
    \propto\ [\bld\beta\mid \alpha_0,\{\bld w^{(a)}\}_{a=1}^r] \ \propto \  (\Gamma(\alpha_0))^r \frac{\prod_{k=1}^K (\beta_{k}^{\frac{g}{K}-1}\ \prod_{a=1}^r(w_{k}^{(a)})^{\alpha_0 \beta_k} )}{\prod_{k=1}^K(\Gamma(\alpha_0\beta_k))^r}.
\end{align*}
The dimension $K$ of the global weights can be potentially large in practice since the distinct global atoms are shared by all the groups, which poses challenges for sampling $\beta$. To tackle that,  we marginalise out  $\alpha_0$ from the joint prior on $(\alpha_0, \beta, \{\bld w\}_{a=1}^r)$, which yields
\begin{align*}
    [\bld\beta, \{\bld w^{(a)}\}_{a=1}^r]\ 
    &\propto\ \int_{0}^{\infty} t^{K-1} \  [\Gamma(t)]^r \frac{\prod_{k=1}^K [e^{-b t\beta_k}(t\beta_{k})^{\frac{g}{K}-1}\ \prod_{a=1}^r(w_{k}^{(a)})^{t \beta_k} ]}{\prod_{k=1}^K[\Gamma(t\beta_k)]^r}\ dt.
\end{align*}
The parameters $\beta$ can be observed to be the normalized version of the random vector $\bld t=(t_1, \ldots, t_{K})$,
\begin{align*}
[\bld t, \{\bld w^{(a)}\}_{a=1}^r]\ &\propto\ t^{K-1} \  \bigg[\Gamma\bigg(\sum_{k=1}^K t_k\bigg)\bigg]^r \frac{\prod_{k=1}^K [e^{-b t_k} t_{k}^{\frac{g}{K}-1}\ \prod_{a=1}^r(w_{k}^{(a)})^{t_k} ]}{\prod_{k=1}^K[\Gamma(t_k)]^r}.
\end{align*}
Note that, the term $[\Gamma(\sum_{k=1}^K t_k)]^r$ above, prevents the posterior density of $\bld t$ from factorizing over $k$. To deal the inconvenience, we observe that 
\begin{align*}
   \bigg[\Gamma(\sum_{k=1}^K t_k)\bigg]^r = \prod_{i=1}^r \bigg[\int_{0}^{\infty} e^{-u_i} u_{i}^{\sum_{k=1}^K t_k - 1} du_i\bigg],
\end{align*}
and introduce a slice using auxiliary random variables $\bld u = (u_1, u_2,\ldots, u_r )$ that yield the  full conditionals:
\begin{align*}
&[\bld u\mid\cdot] \propto \prod_{i=1}^r \mbox{Gamma}\bigg(u_i\bigg| \sum_{k=1}^K t_k, 1\bigg);\quad
[\bld t\mid\cdot] \ \propto\ \prod_{k=1}^K f_k(t_k),
\end{align*}
where
\begin{align*}
& f_k(t) \ \propto\ \frac{1}{[\Gamma(t)]^r} t^{\frac{g}{K}-1} e^{-t(b-\log p_k -\sum_{i=1}^r \log u_i})\ \text{with}\ p_k = \prod_{a=1}^r w^{(a)}_{k}, \ k\in[K].
\end{align*}
We can obtain exact samples from the above log-concave density via a simple rejection sampler equipped with a well-designed covering density \citep{covering}.  Note that, sampling from the full conditional distributions of $\{t_k,\ k\in[K]\}$ are inherently parallelizable. Finally,  we  simply obtain $\beta_k = \frac{t_k}{\sum_{k=1}^K t_k},\ k\in[K]$.

\noindent{\bf Step 3:} To sample from $[\bld w^{(a)}\mid\cdot],\ a\in[r]$ independently, we note that
\begin{align*}
   [\bld w^{(a)}\mid\cdot] \sim \mbox{Dir}(\alpha_0\beta_1 +m^{(a)}_1,\ldots, \alpha_0\beta_K +m^{(a)}_K),
\end{align*}
and set $\bld m^{(a)} = \mbox{rd}(n^{(a)}\times \bld w^{(a)}),\quad  a\in [r].$

\noindent{\bf Step 4.} 
The marginal conditional of clustering indices $[\bld z^{(a)}\mid -], \ a\in[r]$, integrating out population parameters, is
\begin{equation}\label{eq:non_u}
    [\bld z^{(a)}\mid -] \ \propto\ \prod_{k=1}^K \frac{\Gamma_{d}(\nu^{(a)}_{k}/2)\ (\lambda^{(a)}_{0})^{d/2}\ |\Lambda^{(a)}_{0}|^{\nu^{(a)}_{0}/2}}{\Gamma_{d}(\nu^{(a)}_{0}/2)\ (\lambda^{(a)}_{k})^{d/2}\ |\Lambda^{(a)}_{k}|^{\nu^{(a)}_{k}/2}}, \quad \bld z^{(a)} \in \m Z_{N_a,K, \blds m^{(a)}} \quad a\in[r].
\end{equation}
Sampling \eqref{eq:non_u} poses a difficult combinatorial problem and is the most substantial computational bottleneck in our algorithm. We circumnavigate this issue via recasting the problem as a non-uniform sampling task from the space of binary matrices with fixed margin \citep{miller_exact, BRUALDI198033, curveball}, and propose a novel sampling scheme equipped with an MH proposal based on integer-valued optimal transport \citep{Villani2008OptimalTO}.

As an intermediate step, we first develop a computationally convenient MC-EM algorithm \citep{10.2307/1391097}, where instead of sampling from $[\bld z^{(a)}\mid -], \ a\in[r]$, we update the chain with the posterior mode of $[\bld z^{(a)}\mid -], \ a\in[r]$. In particular,  to compute the posterior mode of $[\bld z^{(a)}\mid -], \ a\in[r]$, we go over the following steps.  
\textbf{(i)} First, for $a\in[r]$, we calculate the current component specific means and variances $\bld \mu^{(a)}_{k, \star}, \Sigma^{(a)}_{k, \star}$ for all $a\in[r], k\in[K]$.
\textbf{(ii)} Next,  we  define the $N_a\times K$  cost matrix $\mbox{L}^{(a)}= ((l_{ik})) = \big(\big(-\log \mbox{N}_d(\bld x^{(a)}_{i}\mid \bld\mu^{(a)}_{k,\star}, \Sigma^{(a)}_{k, \star}))$, column sum vectors $\bld c^{(a)}=\bld m^{(a)}$ and row sum  vectors $\bld r^{(a)} = \bld 1_{N_a}$, where $\bld 1_{N_a}$ is a vector of $N_a$ $1$s.
\textbf{(iii)} Next, given the two vectors $\bld r^{(a)}, \bld c^{(a)}$, we define the polytope of $K\times N_a$ binary cluster membership matrices $\mbox{U}(\bld r^{(a)}, \bld c^{(a)}):=\{\mbox{B}\mid \mbox{B}\bld 1_{N_a} = \bld r^{(a)};\ \mbox{B}^{\T}1_{K} = \bld c^{(a)}\}$, and solve the constrained binary optimal transport problem \citep{Villani2008OptimalTO}
$\mbox{B}^{(a)}= \mbox{\rm argmin}_{\mbox{B}\in \mbox{U}(\bld r^{(a)},\ \bld c^{(a)})} \langle\mbox{B}, \mbox{L}^{(a)} \rangle$ where $\langle\mbox{B}, \mbox{L}^{(a)}\rangle = \rm{tr}(\mbox{B}^\T \mbox{L}^{(a)})$. Finally, we obtain the modal clustering indices $\bld z^{(a)}$ from the binary cluster membership matrices $\mbox{B}^{(a)}, a\in[r]$, completing the MC-EM algorithm. 

\subsection*{Posterior Summary}
Suppose $S_{T} = \{s_1,\ldots, s_T\}$ be $T$ post burn-in draws from the marginal posterior of  clustering distribution. For each clustering configuration $s\in S_{T}$, one can obtain the association matrix $\eta(s)$ of order $N\times N$, whose $(i,j)$-th element is an indicator whether the $i$-th and the $j$-th observation are clustered together or not. Element wise averaging of these $T$ association matrices yield the pairwise clustering probability matrix, denoted by $\bar{\eta} = \frac{1}{T}\sum_{s\in S_{T}} \eta(s) $. To summarize the MCMC draws, we  adopt the least square model-based clustering introduced in \citet{Dahl2006} to obtain
$s_{\rm LS} = \mbox{\rm argmin}_{s\in S_{T}}\sum_{i=1}^{N} \sum_{j=1}^{N} (\eta_{ij}(s) - \bar{\eta}_{ij})^2.$
Since $s_{\rm LS}\in S_{T}$ by construction, the notion of balance  is retained in the resulting $s_{\rm LS}$.

\RestyleAlgo{ruled}

\SetKwComment{Comment}{/* }{ */}

\begin{algorithm}[hbt!]
\caption{Weighted rectangular loop algorithm}\label{algo:weighted_rec_loop}
\textbf{Input:} An initial binary matrix $A_0$, and total number of iterations $T$.\\
\For{$t=1,\ldots, T$}{
\Comment*[r]{Try to find a $2\times2$ checker-board.}
Choose one row and one column $(r_1,c_1)$ uniformly at random.\\
  \eIf{$A_{t-1}(r_1,c_1) = 1$}{
   Choose one column $c_2$ at random among all the 0 entries in $r_1$.\\
   Choose one row $r_2$ at random among all the 1 entries in $c_2$.\\
  }
  {
  Choose one row $r_2$ at random among all the 1 entries in $c_1$.\\
  Choose one column $c_2$ at random among all the 0 entries in $r_2$.\\
    }
    
\Comment*[r]{If we find a checker-board, we may want to  swap.}
  
  \eIf{The sub-matrix extracted from $r_1, r_2, c_1, c_2$ is a checkerboard unit}{
  Obtain $B_{t}$ from $A_{t-1}$ by swapping the checker-board.\\
  Calculate $p_{t}=\frac{P(B_{t})}{P(B_{t}) + P(A_{t-1})}$.\\
  Draw $r_t\sim \mbox{Bernoulli}(p_{t})$.\\
  \eIf{$r_t = 1$}{
  Set $A_{t}=B_{t}$.
  }
  { Set $A_{t}=A_{t-1}$
}
}
  {$A_{t}=A_{t-1}$
}
}
\end{algorithm}
\subsection{Weighted Rectangular Loop Algorithm (W-RLA) and Proof of Theorem 2 in the section 4 of the main document}\label{sup_wrla}
Given fixed row sums $\mathbf{r} = (r_1,\ldots, r_u)^{\T}$ and column sums $\mathbf{c} = (c_1,\ldots, c_v)^{\T}$, we denote the collection of all $u\times v $ binary matrices by $\mbox{U}(\mathbf{r}, \mathbf{c})$. Further, we denote by $\Omega=(\omega_{ij})\in [0,\infty)$ a  non negative weight matrix representing the relative probability of observing a count of $1$ at the $(i,j)$-th cell. Then the likelihood associated with the observed binary matrix $H\in\mbox{U}(\mathbf{r}, \mathbf{c})$ as
$\mbox{P}(H) = (1/\kappa) \prod_{i,j} \omega_{ij}^{h_{ij}},\ \kappa = \sum_{H\in\mbox{U}(\mathbf{r}, \mathbf{c})}\ \prod_{i,j} \omega_{ij}^{h_{ij}}$. 
Let  $\mbox{U}^{\prime}(\mathbf{r}, \mathbf{c}) = \{H\in\mbox{U}(\mathbf{r}, \mathbf{c}):\ P(H)>0\}$ denote the subset of matrices in $\mbox{U}(\mathbf{r}, \mathbf{c})$ with positive probability. Then, for  $H_1,H_2\in\mbox{U}^{\prime}(\mathbf{r}, \mathbf{c})$,  the relative probability of the two observed matrices is
$\mbox{P}(H_1)/\mbox{P}(H_2) = \prod_{\{i,j: h_{1, ij}=1, h_{2, ij}=0\}} \ \omega_{ij}^{h_{1, ij}}/ \prod_{\{i,j: h_{1, ij}=0, h_{2, ij}=1\}} \   \omega_{ij}^{h_{2, ij}}$.
With these notations, we are all set to introduce a \emph{Weighted Rectangular Loop Algorithm} (W-RLA)  for non-uniform sampling from the space of fixed margin binary matrices $H\in\mbox{U}(\mathbf{r}, \mathbf{c})$, given the weight matrix $\Omega=(\omega_{ij})\in [0,\infty)$.  To that end, let us first record that the identity matrix of order $2$, and the $2\times2$ matrix  with all zero  diagonal entries and all one off-diagonal entries, are referred to as \emph{checker-board} matrices. 

The Theorem 2 in the main document states that, given an weight matrix $\Omega$ with $\omega_{ij} > 0$, and fixed row-sum and column sum vectors $(\bld r, \bld c)$,  the Weighted Rectangular Loop Algorithm (W-RLA) generates a
Markov chain with stationary distribution given by $\mbox{P}(H) = (1/\kappa) \prod_{i,j} \omega_{ij}^{h_{ij}},\ \kappa = \sum_{H\in\mbox{U}(\mathbf{r}, \mathbf{c})}\ \prod_{i,j} \omega_{ij}^{h_{ij}}$.

\noindent{\bf Proof of Theorem 2.}
Combine proof of the Theorem 1 in \citet{fout2020nonuniform}  and proof of Theorem 2 in \cite{recloop} to get the result. 

\section{Extensions}\label{extensions}
As we mentioned in the main document, our approach enjoys seamless adaption  to include key extensions present in  the fair clustering literature \citep{NIPS2017_978fce5b, https://doi.org/10.48550/arxiv.2002.07892, NEURIPS2020_95f2b84d}; and inherits the usual benefits of model based clustering, e.g,  simultaneous learning of the number of clusters and the cluster memberships \citep{doi:10.1198/016214502760047131, 10.1093/biomet/asac051, doi:10.1080/01621459.2016.1255636};
providing a framework to potentially handle an enormous range of pressing practical  complexities --  non-isotropic covariances within clusters, mixed data type, censoring, missing values, variable selection \citep{complicated}, covariate adjustment, etc. Two specific examples are in order.
\subsection*{Imperfect Knowledge on the Protected Attribute}
Much of the prior work on fair clustering assumes complete knowledge of group membership.  \citet{NEURIPS2020_95f2b84d} generalized this by assuming imperfect knowledge of group membership through probabilistic assignments, where it  is assumed that, for individual $i\in[N]$, the protected label $a_{i} = a$ with probability $p^{(a)}_{i}$ (known) such that $\sum_{a=1}^r p^{(a)}_{i} = 1, \ i\in[N]$. The goal is to balance the \emph{expected color} in each cluster. Our fully probabilistic set up enables us seamlessly achieve this via augmenting every cycle of our computational scheme by an additional sampling step as follows $a_i = a\quad \text{with probability}\ p^{(a)}_{i},\ a\in[r],\ i\in[N]$.

\subsection*{Presence of Mixed Variable Types or Censoring}
Model-based approaches often enables us handle practical  complexities \citep{complicated} e.g mixed data type, censoring, missing values, in a seamless manner. For simplicity in exposition, we shall present the case with mixed data type where some variables are ordinal, and rest are continuous and potentially censored.  Recall that, we have observed data  $\{(\bld x_i, a_i)\}_{i=1}^N$ with $(\bld x_i, a_i) \in \m X \times \m A$, where $\bld x_i$ denotes the $d$-variate observation for the $i$th data unit, and $a_i$ the corresponding level of the protected attribute. Here, we shall further assume $\m X = \m X_{C}\times \m X_{D}$, where $\m X_{C}$ and $\m X_{D}$ are the supports of the potentially censored continuous variables and ordinal discrete variables respectively. We  introduce $d$-variate Gaussian latent variables $\{u_i\}_{i=1}^N$ to operate in this set-up:
\begin{align*}
x_{ij} = 
    \begin{cases}
    \sum_{l=1}^{L_j}\ d_{j,l}\ \mathbf{1}(a_{j,l-1} < u_{ij} \leq a_{j,l} ),\ \text{if}\quad \mbox{support}(x_{ij})\subset\m X_{D}, \\
    u_{ij}\ \mathbf{1}(b_{j} \leq u_{ij} \leq c_{j} ) + b_j\ \mathbf{1}(u_{ij}<b_j) + c_j\ \mathbf{1}(u_{ij}>c_j), \ \text{if}\quad \mbox{support}(x_{ij})\subset\m X_{C},
    \end{cases}
\end{align*}
where $i\in[N],\ j\in[d],\ \mathbf{1}(\cdot)$ is the indicator function. It is straightforward to use the latent Gaussian variable approach to allow for unordered categorical variables \citep{https://doi.org/10.2307/2290350}, is omitted here to avoid redundancies.

\section{Performance metrics (Fair-score in section 2.1 in the main document)}\label{performance}
The Fair-Score introduced in equation (3) of section (2.1) in the main document, serves as a criterion towards a principled assessment of the clustering performance as it strikes a balance between posterior maximization while maintaining fairness. Here, we provide further numerical evidence on the the validity of the fair-score in our context.
\subsection*{Example 1 (Well-specified case).}
\begin{figure}
\begin{center}
\includegraphics[width=16cm, height =4cm]{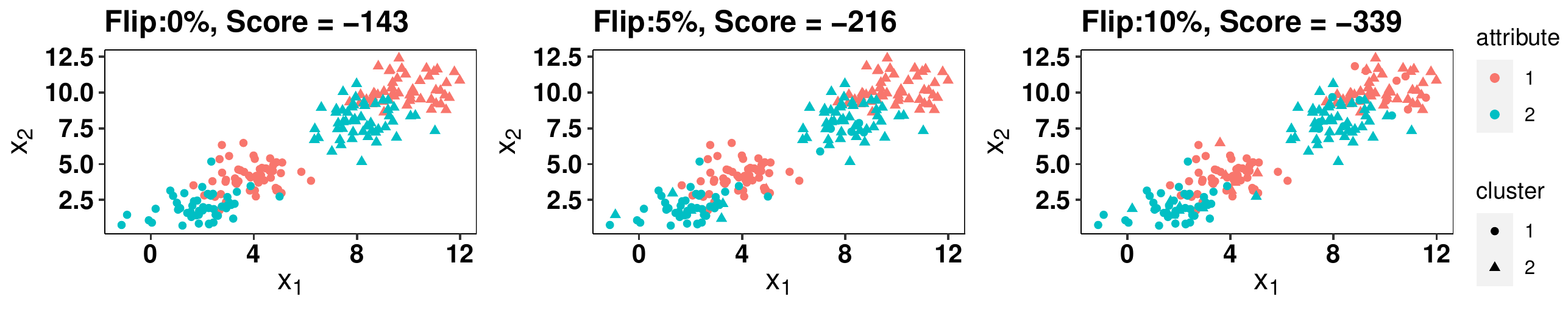} 
\caption{\textbf{Performance evaluation (well-specified case).} The \textbf{left panel} presents the true clustering configuration that yields fair-score equal to $-142.85$, and balance equal to $0.96$. The \textbf{middle} and the \textbf{right} panel present the clustering configurations and corresponding fair-score, where the configurations are obtained via flipping $5\%$ and $10\%$ of the true clustering indices respectively. The balance in the clusters
remain the same, but the criteria steadily decreases with increase in flipping proportion $p$, as expected. }\label{fig:criteria}
\end{center}
\end{figure}

We generate data from the model in equation (4)-(6) with the value of the hyper-parameters $b=0.1, g = 100$, which  results in approximately balanced  clusters. We fix the true number of clusters $K_{true}=2$, and the sample sizes $N^{(1)}=N^{(2)}=100$. We consider the  following case: in the first cluster, the individuals with $a=1$ are generated from $N_2((4,4)^{\prime}, S)$ and individuals with $a=2$ are generated from $N_2((2,2)^{\prime}, S)$.  In the second cluster,individuals with $a=1$ are generated from $N_2((10,10)^{\prime}, S)$ and  individuals with $a=2$ are generated from $N_2((8,8)^{\prime}, S)$ where $S = 3\times[\rho 11^{\T} + (1-\rho)I_2]$ with $\rho = 0.3$. In this well separated example, the true clustering configuration yields the value of the criteria in fair-score equal to $-142.85$, and balance equal to $0.96$. Next, we randomly flip the clustering indices corresponding to $100\times p\%$  of the observations. Consequently, the balance in the clusters
remain the same, but the value of the criteria steadily decrease with the increase in $p$, refer to Figure \ref{fig:criteria} for details. This empirically justifies the usage of the  fair-score to assess the clustering performance, while maintaining balance.

\subsection*{Example 2 (Mis-specified case).} 
\begin{figure}
\begin{center}
\includegraphics[width=16cm, height =4cm]{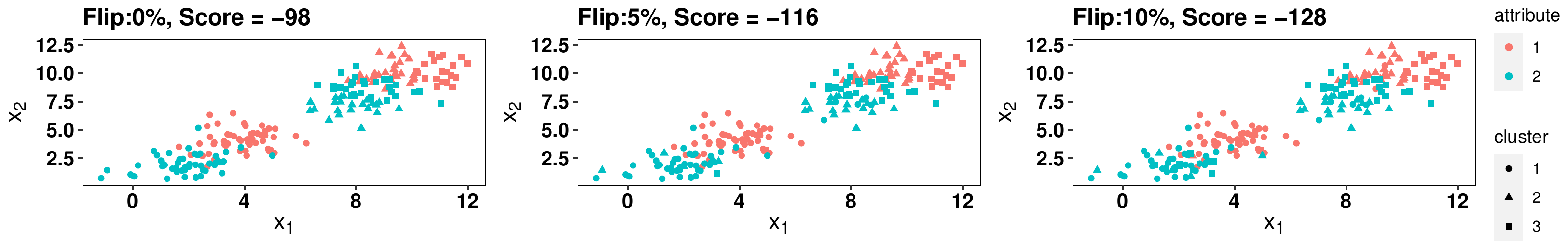} 
\caption{\textbf{Performance evaluation (mis-specified case).} Here, we deliberately misspecify the number of clusters, and set $K=3$ instead of $K=K_{true}=2$. The \textbf{left panel} presents a intuitively plausible clustering configuration that yields the fair-score equal to $-96.13$, and balance equal to $0.85$. The \textbf{middle} and the \textbf{right} panel present the clustering configurations and corresponding fair-score, where the configurations are obtained via flipping $5\%$ and $10\%$ of the initial clustering indices respectively. The balance in the clusters
remain the same, but the criteria steadily decreases with increase in flipping proportion $p$, as expected. }\label{fig:criteria_miss}
\end{center}
\end{figure}

We generate the data set in the same manner as Example 1, but deliberately mis-specify the number of clusters, and set $K=3$ instead of $K=K_{true}=2$.  In this example, a intuitively plausible clustering configuration yields fair-score equal to $-93.19$, and balance equal to $0.85$. Noteworthy, since the balance condition is relaxed in Example 2 compared to Example 1, the fair-scores are not directly comparable across the examples.
Next, as earlier, we randomly flip the clustering indices corresponding to $100\times p\%$  of the observations. Consequently, the balance in the clusters
remain the same, but the value of the criteria steadily decrease with the increase in $p$, refer to Figure \ref{fig:criteria_miss} for details. This empirically justifies the usage of the fair-score to assess the clustering performance while maintaining balance, even under model mis-specification.

\begin{figure}[!htbp]
\begin{center}
\includegraphics[width=12cm, height =18cm]{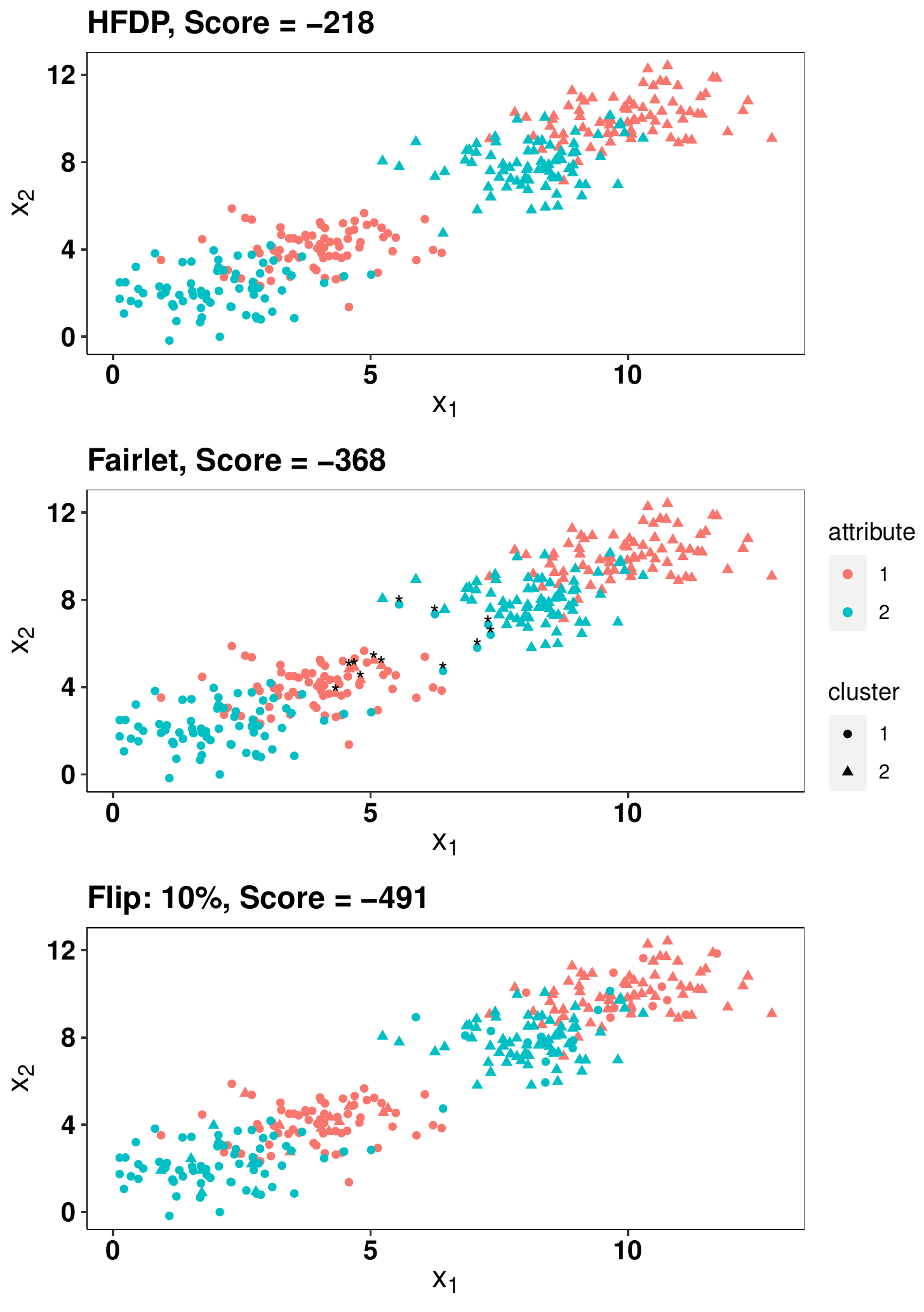} 
\centering{{\caption{\textbf{Figure 3 in the main document (zoomed in). }\emph{For one instance in the simulation set up (A.1), the left and middle panel present clustering configuration via HFDP and  fair k-means\citep{NIPS2017_978fce5b}, respectively. The differences between the two configurations are marked with black "*". Non-optimal exchange of points in fair k-means leads to substantial decrease in fair-score. The right panel presents the clustering configurations obtained via randomly flipping $10\%$ of the MAP of the HFDP clustering indices, to demonstrate that fair-score is low for non-sensical clustering configurations.}}\label{fig:fair_score_validity}}}.
\end{center}
\end{figure}

\end{document}